\documentclass{article}

\usepackage{arxiv}

\newcommand{\paperrepo}{https://github.com/Peregrine123/ROPD_official}

\usepackage{amsmath,amsfonts,bm}

\usepackage{amsmath,amsfonts,bm}

\def\eqref#1{equation~\ref{#1}}

\def\1{\bm{1}}

\DeclareMathAlphabet{\mathsfit}{\encodingdefault}{\sfdefault}{m}{sl}
\SetMathAlphabet{\mathsfit}{bold}{\encodingdefault}{\sfdefault}{bx}{n}

\usepackage[table]{xcolor}
\usepackage[utf8]{inputenc}
\usepackage[T1]{fontenc}
\usepackage{url}
\usepackage[round,authoryear]{natbib}
\setcitestyle{authoryear,round,citesep={;},aysep={,},yysep={;}}
\usepackage{booktabs}
\usepackage{amsfonts}
\usepackage{nicefrac}
\usepackage{microtype}
\usepackage{tabularx}
\usepackage{siunitx}
\usepackage{multirow}
\usepackage{capt-of}
\usepackage{graphicx}
\usepackage{tcolorbox}
\usepackage{fvextra}
\DefineVerbatimEnvironment{PromptVerbatim}{Verbatim}{breaklines=true, fontsize=\scriptsize}
\tcbuselibrary{breakable}
\usepackage{makecell}
\usepackage{enumitem}
\usepackage{placeins}
\usepackage{tikz}

\usepackage{amsmath}
\usepackage{amssymb}
\usepackage{mathtools}
\usepackage{amsthm}

\usepackage{algorithm}
\usepackage{algpseudocode}

\definecolor{myred}{rgb}{0.8, 0.2, 0.2}
\definecolor{myblue}{rgb}{0.2, 0.2, 0.8}
\definecolor{TableTitleBg}{RGB}{63,63,63}
\definecolor{LightGrayBg}{RGB}{242,242,242}
\definecolor{LLMHeaderBg}{RGB}{220,220,220}
\definecolor{WhiteText}{RGB}{255,255,255}
\definecolor{BlackText}{RGB}{0,0,0}
\definecolor{RedText}{RGB}{255,0,0}

\newtcolorbox{keyinsightbox}[1][]{%
  breakable,
  colback=LightGrayBg!30,
  colframe=LLMHeaderBg!80!black,
  colbacktitle=LLMHeaderBg,
  coltitle=BlackText,
  fonttitle=\bfseries\small,
  fontupper=\small,
  boxrule=0.8pt,
  arc=2pt,
  left=8pt,
  right=8pt,
  top=3pt,
  bottom=3pt,
  before skip=4pt,
  after skip=6pt,
  #1,
}

\AtBeginDocument{%
  \setlist{itemsep=2pt, parsep=0pt, topsep=4pt, partopsep=0pt}
}

\definecolor{LeaderboardRowBlue}{RGB}{227,241,248}
\definecolor{LeaderboardGainGreen}{RGB}{0,150,80}
\definecolor{LeaderboardLossRed}{RGB}{190,45,65}

\colorlet{BOPDRowBlue}{LeaderboardRowBlue}
\colorlet{BOPDGainGreen}{LeaderboardGainGreen}
\colorlet{BOPDLossRed}{LeaderboardLossRed}

\newcommand{\gain}[1]{\textcolor{LeaderboardGainGreen}{#1}}
\newcommand{\loss}[1]{\textcolor{LeaderboardLossRed}{#1}}
\makeatletter
\newcounter{scoreannot}
\newcommand{\scoreannotoverlays}{}
\newcommand{\scoreannot@add}[2]{%
  \g@addto@macro\scoreannotoverlays{%
    \node[
      anchor=mid west,
      inner sep=0pt,
      outer sep=0pt,
      xshift=0.10em,
      yshift=-0.10ex,
      font=\fontsize{6}{6}\selectfont
    ] at (#1.base east) {#2};%
  }%
}
\newif\ifscoreinline
\newcommand{\scoreannot}[2]{%
  \ifscoreinline
    #1\,{\fontsize{6}{6}\selectfont #2}%
  \else
    \scoreannot@overlay{#1}{#2}%
  \fi
}
\newcommand{\scoreannot@overlay}[2]{%
  \stepcounter{scoreannot}%
  \edef\scoreannotid{scoreannot\arabic{scoreannot}}%
  \tikz[remember picture,baseline=(\scoreannotid.base)]{%
    \node[inner sep=0pt, outer sep=0pt] (\scoreannotid) {#1};%
  }%
  \expandafter\scoreannot@add\expandafter{\scoreannotid}{#2}%
}
\newcommand{\PrintScoreAnnotations}{%
  \tikz[remember picture,overlay]{\scoreannotoverlays}%
  \global\let\scoreannotoverlays\@empty
}
\makeatother
\newcommand{\scoreup}[2]{\scoreannot{#1}{\gain{#2}}}
\newcommand{\scoredown}[2]{\scoreannot{#1}{\loss{#2}}}
\newcommand{\bestup}[2]{\scoreup{\textbf{#1}}{#2}}

\newcommand{\secondup}[2]{\scoreup{\underline{#1}}{#2}}

\usepackage{hyperref}
\hypersetup{
  colorlinks=true,
  linkcolor=blue!55!black,
  citecolor=blue!55!black,
  urlcolor=blue!55!black,
  pdfborder={0 0 0},
}

\title{Rethinking Black-Box On-Policy Distillation:\\ A Rubric-based Analysis}

\renewcommand{\shorttitle}{Rethinking Black-Box On-Policy Distillation}

\author{%
Zhepei Hong$^{1}$ \quad Yukai Wang$^{2}$ \quad Mao Zheng$^{2}$ \quad Mingyang Song$^{2}$ \quad Gengsheng Li$^{2}$ \\
Houcheng Jiang$^{1}$ \quad Dan Zhang$^{1}$ \quad Haiyun Guo$^{1}$ \quad Junfeng Fang$^{1\dagger}$ \quad Tat-Seng Chua$^{1}$ \\[8pt]
{\normalsize $^{1}$National University of Singapore} \\[2pt]
{\normalsize $^{2}$Tencent} \\[8pt]
{\small $^{\dagger}$Corresponding author} \\[2pt]
{\small \texttt{hongzhepei@gmail.com}, \texttt{fangjf1997@gmail.com}}
}

\date{}

\begin{document}

\maketitle

\begin{abstract}
On-policy distillation (OPD) is a powerful paradigm for LLM post-training, yet its reliance on teacher logits restricts it to white-box teachers. In this paper, we study whether rubric-based supervision can provide a text-only feedback interface for black-box OPD, and what makes it effective. Our instantiation, Rubric-based On-Policy Distillation (ROPD), induces rubrics by contrasting teacher responses with the student's current rollouts and uses them to score these rollouts for on-policy optimization.
We find that, under matched settings, text-only ROPD outperforms logit-based OPD methods on four competition-math benchmarks, reaching the best AIME24 accuracy of standard logit-based OPD with 9.6$\times$ fewer training samples.
Our analysis yields three findings: rubric rewards track correctness far better than teacher log-likelihood (AUC 0.90 vs.\ 0.35); training with them repairs more failed criteria and regresses fewer passed ones than logit-based supervision; and contrasting teacher responses with student rollouts is central to rubric quality.
These results position rubric-based OPD as a flexible, black-box-compatible alternative to logit-based OPD. Our implementation is available at \url{\paperrepo}.

\end{abstract}

\begin{figure}[!ht]
\centering
\includegraphics[width=\textwidth]{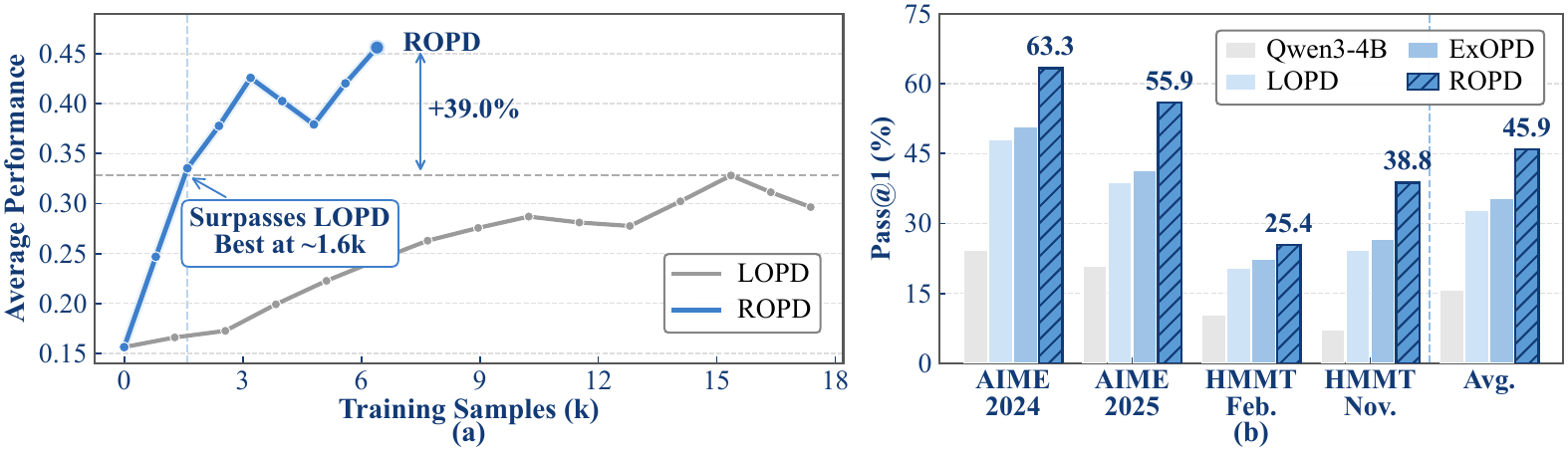}
\vspace{-5mm}
\caption{%
  \textbf{ROPD efficiency and reasoning performance.} (a) Training dynamics averaged over four math benchmarks (AIME 24/25 and HMMT 25 Feb./Nov.~\citep{aime2024,aime2025,hmmt2025}). ROPD surpasses LOPD's best accuracy early in training (Section~\ref{subsec:whitebox}). (b) Comparative results under matched settings (DAPO-Math-17K training set, Qwen3-4B student, Qwen3-30B-A3B teacher~\citep{yu2025dapo,yang2025qwen3technicalreport}). See Section~\ref{sec:setup} for details.
}
\label{fig:teaser}
\end{figure}

\section{Introduction}

On-policy distillation (OPD) trains a student on feedback computed over its own rollouts, keeping the training signal aligned with the current policy \citep{agarwal2024gkd,lu2025onpolicydistillation,song2026surveyopd}. By using teacher logits as a dense signal \citep{gu2024minillm}, OPD has been widely adopted in recent open-source reasoning models \citep{yang2025qwen3technicalreport,xiao2026mimo,deepseekai2026deepseekv4}. However, this requires white-box teacher access; as frontier models increasingly expose only text through APIs, the set of usable OPD teachers is restricted.

Several lines of work explore text-only supervision for black-box distillation, including preference feedback \citep{singh2025orpodistill}, adversarial discriminators \citep{ye2026gad}, LLM-as-a-Judge scoring \citep{shen2026rljudgekd}, and fine-tuning on teacher demonstrations \citep{kim2016sequence}. Some of these methods, such as GAD \citep{ye2026gad} and OVD \citep{xiong2026ovd}, update the student on its own current rollouts, showing that on-policy distillation does not depend on access to teacher logits. The remaining limitation lies in the feedback: these methods compress each response into a single preference or score, which indicates how good a rollout is but not which properties it lacks. This raises our central question: \textit{can text-only teacher feedback make black-box on-policy distillation competitive with logit-based OPD, and what makes it effective?}

Answering this question requires turning teacher responses into effective feedback for the current student. A black-box teacher's responses carry knowledge about problem-solving strategies and answer quality; useful feedback should be teacher-grounded, drawing on this knowledge, and student-conditioned, targeting gaps in the student's own rollouts. Rubrics offer a natural interface: they decompose quality into explicit, self-contained criteria that are evaluated separately and aggregated into a weighted reward. Because each criterion can encode a teacher-demonstrated property while being checked against the student's own response, rubrics connect teacher knowledge to current student gaps.

We investigate this question through \textbf{ROPD (Rubric-based On-Policy Distillation)}, a concrete instantiation of rubric-based supervision for black-box on-policy distillation. Applying rubrics directly in this setting faces three difficulties: (i) criteria derived only from teacher answers describe desirable properties without revealing what the current student lacks; (ii) criteria from a single teacher answer inherit path-specific choices and may penalize other valid solutions; (iii) rubrics constructed separately for each rollout introduce score differences unrelated to response quality, so rollouts of the same prompt should be judged on a common basis.

ROPD addresses these difficulties through \textit{teacher--student contrastive rubric induction}. For each prompt, a \textit{Rubricator} jointly examines multiple teacher responses and the current student rollout group, and induces one set of weighted, self-contained criteria shared by all rollouts. A \textit{Verifier} evaluates each rollout against these criteria without response provenance, and GRPO~\citep{shao2024deepseekmath} uses the aggregated verdicts as rewards to update the policy that generated the rollouts, preserving the on-policy loop under text-only teacher access. Unlike rubric-based RL, whose rubrics define the target quality of a task, ROPD derives criteria from the teacher--student contrast to transfer a teacher's competence to the current student (Section~\ref{sec:related}).

Under matched teacher, student, data, and training configurations, text-only ROPD outperforms logit-based OPD methods on all four math benchmarks and reaches the best AIME24 accuracy of standard logit-based OPD with 9.6$\times$ fewer training samples. Our analysis further explains this result through three findings on reward quality, training dynamics, and rubric construction, which together indicate that rubric-based supervision can serve as an effective feedback interface for black-box OPD. To our knowledge, this is the first systematic study of rubric-based supervision for black-box on-policy distillation.

Our main contributions are as follows:
\begin{itemize}[leftmargin=*,itemsep=4pt,parsep=0pt]
    \item \textbf{Rubrics as a feedback interface for black-box on-policy distillation.} We study whether rubric-based supervision can stand in for teacher logits under text-only access; under matched settings, ROPD outperforms logit-based OPD on all four math benchmarks.

    \item \textbf{Insight 1: rubric rewards track correctness, whereas teacher likelihood does not.} On a fixed rollout pool, rubric rewards separate correct from incorrect rollouts far better than teacher log-likelihood (AUC $0.90$ vs.\ $0.35$), consistent with teacher likelihood favoring fluent but flawed reasoning over correct but stylistically divergent solutions.

    \item \textbf{Insight 2: under matched training, rubric rewards stabilize criterion-level progress relative to logit-based supervision.} During training, ROPD repairs more failed criteria and regresses fewer previously passed criteria than LOPD, indicating that decomposed rewards support targeted progress without eroding acquired capabilities.

    \item \textbf{Insight 3: teacher--student contrast is central to rubric quality.} Controlled source and component ablations indicate that constructing rubrics from both teacher responses and current student rollouts is central to reward quality: multi-teacher coverage is the dominant factor, while shared rubrics and blind verification provide complementary gains in the evaluated setting.
\end{itemize}

\section{Rubric-based On-policy Distillation}
\label{sec:method}

ROPD instantiates rubric-based supervision for black-box on-policy distillation by making \textbf{rubric construction part of the feedback loop}. Teacher responses establish which solution properties are desirable, while the current student's rollouts reveal where those properties are missing. ROPD brings these two sources of evidence together to construct a shared evaluation standard, then converts criterion-wise judgments into rewards for updating the student. The resulting supervision is grounded in teacher competence and conditioned on the current policy's behavior.

\begin{figure*}[t]
\centering
\includegraphics[width=0.98\textwidth]{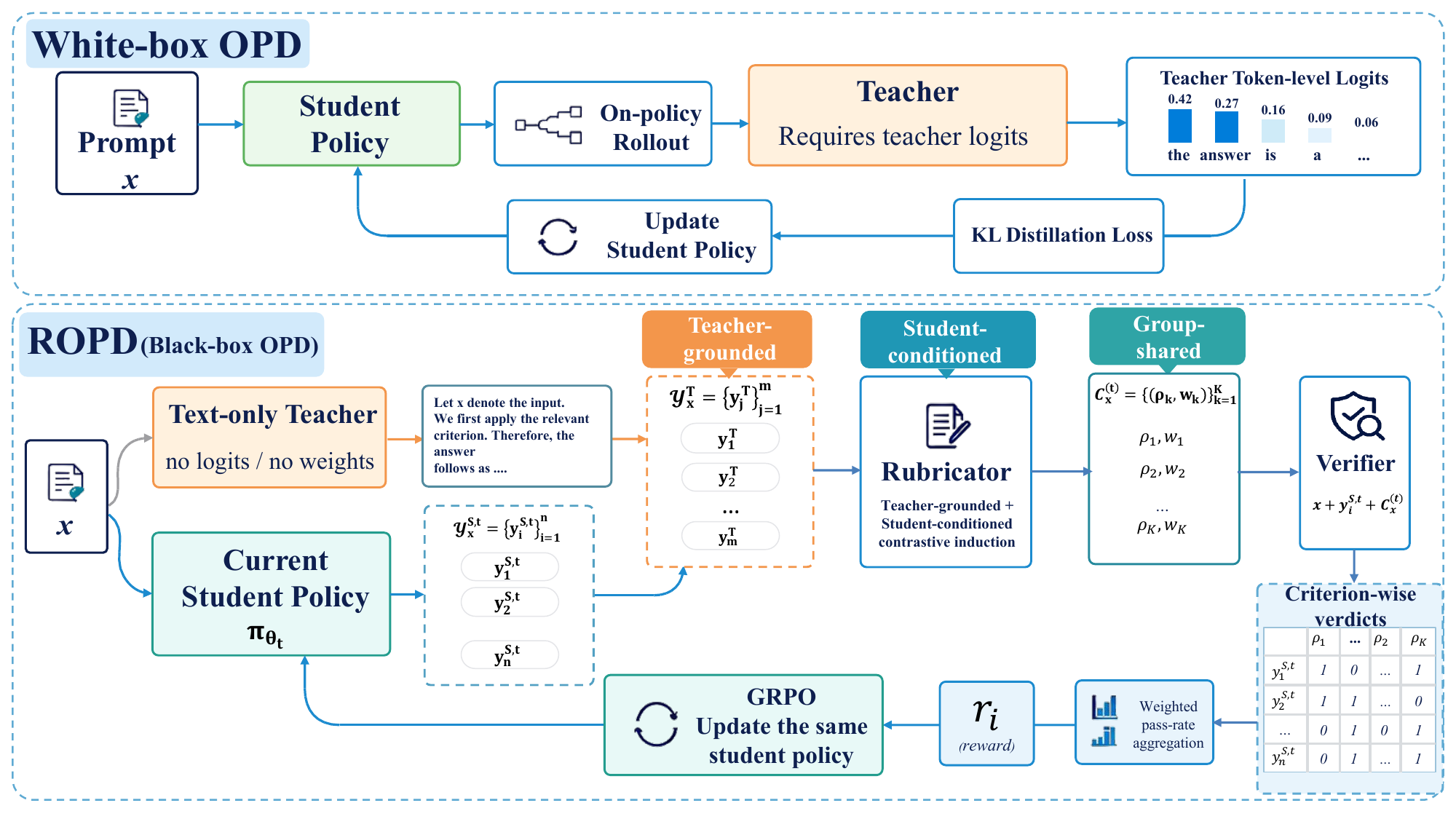}
\caption{\textbf{White-box OPD vs. ROPD.} Top: White-box OPD uses teacher logits for token-level KL supervision. Bottom: ROPD constructs teacher-student contrastive rubrics from text-only access, converting criterion-wise verification into rewards. Both methods use on-policy student rollouts.}
\label{fig:1}
\vspace{-5pt}
\end{figure*}

Conditioning rubric induction on student rollouts makes current learning gaps visible; using multiple teacher responses provides a broader basis for identifying valid solution properties; and sharing the resulting rubric establishes a consistent basis for reward comparison. As illustrated in Figure~\ref{fig:1}, ROPD implements these requirements through a \textit{Rubricator} for teacher--student contrastive rubric induction and a \textit{Verifier} for evaluating individual responses against the induced criteria.

\subsection{Black-box On-policy Distillation}
\label{sec:setting}


On-policy distillation trains a student model using teacher-derived feedback on the student's own rollouts~\citep{agarwal2024gkd,song2026surveyopd}. We consider a black-box teacher $\pi_T$ accessible only through text outputs. For a prompt $x$, the teacher provides a response set $\mathcal{Y}^T_x=\{y^T_j\}_{j=1}^{m}$, and the current student policy $\pi_{\theta_t}$ samples a rollout group $\mathcal{Y}^{S,t}_x=\{y^{S,t}_i\}_{i=1}^{n}$. Teacher token probabilities, hidden states, and model weights are unavailable.


This setting imposes two requirements on the supervision signal. It must recover useful guidance from teacher text, and it must evaluate responses generated by the \textit{current student policy}. The latter distinguishes black-box on-policy distillation from supervised fine-tuning on teacher demonstrations~\citep{kim2016sequence} and preference optimization on fixed teacher-generated pairs~\citep{singh2025orpodistill}, which do not necessarily involve the current student in feedback construction. Here, the student generates rollouts, receives teacher-grounded feedback on them, and is updated from it.




\subsection{Teacher-grounded Contrastive Rubric Induction}
\label{sec:rubric_induction}

The teacher response set provides evidence about desirable solution behavior, but does not by itself specify how to score the student's answers. Useful feedback therefore requires extracting criteria from teacher demonstrations and applying them to the student's own responses.

Recent rubric-based RL methods decompose response quality into reference-grounded criteria and aggregate criterion-level evaluations into rewards~\citep{gunjal2025rar,li2026rubricarm}. ROPD retains this reward structure but \textbf{conditions the criteria themselves on the teacher--student gap}: the current student's responses help construct the standard used to evaluate them.

A rubric derived only from teacher answers can describe a high-quality solution, yet remain insensitive to the student's current error distribution. This matters because group-relative optimization depends on reward differences among student rollouts: a criterion that every rollout passes, or every rollout fails, yields no learning signal.



For each prompt $x$ at step $t$, the Rubricator jointly processes the teacher responses and student rollouts:
\begin{equation}
\mathcal{C}^{(t)}_x
=
\mathrm{Rubricator}
\big(x,\mathcal{Y}^T_x,\mathcal{Y}^{S,t}_x\big)
=
\{(\rho_k,w_k)\}_{k=1}^{K},
\end{equation}
where each $\rho_k$ is a self-contained textual criterion, and $w_k>0$ reflects its importance in the current teacher--student gap. Each criterion must be evaluable given only the prompt and a single response.

This construction assigns complementary roles to the two response sets: \textit{teacher answers ground the quality standards; student rollouts inform which shortcomings those standards should address}. The resulting rubric is therefore both \textit{teacher-grounded} and \textit{student-conditioned}. Its dependence on $\mathcal{Y}^{S,t}_x$ lets the criteria track the student's behavior as training proceeds.


\textbf{Multiple teacher responses and a shared rubric.}
The Rubricator receives $m>1$ teacher responses per prompt, so that criteria can capture solution properties shared across valid reasoning paths rather than choices specific to a single demonstration. All student rollouts for the same prompt are evaluated using the same $\mathcal{C}^{(t)}_x$, including both its criteria and weights. Sharing the rubric holds the evaluation standard fixed within the group. Section~\ref{sec:ablation} ablates these design choices.

\subsection{Criterion-wise Verification and On-policy Reward}
\label{sec:verification}

The induced rubric specifies what should be evaluated; verification checks whether each response meets it. ROPD separates these roles to preserve a clear information boundary: \textbf{teacher--student comparison shapes the criteria, while each response is judged against the shared standard}.

\textbf{Blind verification of explicit criteria.}
Because each criterion is self-contained, the Verifier requires only the prompt, the response, and the criterion. The Verifier sees neither teacher responses nor the source of a rollout; this \textit{blind, criterion-wise evaluation} reduces source-dependent bias:
\begin{equation}
    v_{i,k}
    =
    \mathrm{Verifier}
    \big(
        x,\;
        y^{S,t}_{i},\;
        \rho_k
    \big),
    \qquad
    v_{i,k}\in\{0,1\},
\end{equation}
where $v_{i,k}=1$ indicates that rollout $y^{S,t}_i$ satisfies criterion $\rho_k$.



\textbf{From criterion satisfaction to comparable rewards.}
The binary verdicts are aggregated into a scalar reward through their weighted pass rate:
\begin{equation}
    r_i
    =
    \frac{
        \sum_{k=1}^{K} w_k\,v_{i,k}
    }{
        \sum_{k=1}^{K} w_k+\epsilon
    },
\end{equation}
where $\epsilon$ is a small constant for numerical stability. The shared weights provide a common aggregation rule across rollouts, and the criterion-level verdicts keep each scalar reward interpretable.



\textbf{Closing the on-policy feedback loop.}
Finally, GRPO~\citep{shao2024deepseekmath} uses $\{r_i\}_{i=1}^{n}$ to update $\pi_{\theta_t}$, the policy that generated the rollout group. The complete feedback loop is thus \textit{current student rollouts $\rightarrow$ teacher--student contrastive rubric $\rightarrow$ criterion-wise rewards $\rightarrow$ student policy update}. In our experiments, the teacher model serves as both the Rubricator and the Verifier. The complete training algorithm is provided in Appendix~\ref{app:algorithm}.


\section{Main Results}
\label{sec:experiments}

\subsection{Setup}
\label{sec:setup}
\textbf{Models.} We use Qwen3-4B~\citep{yang2025qwen3technicalreport} as the student. In the black-box setting, GPT-5.2-chat-latest~\citep{openai2025gpt52} is accessed through text-only API responses and is compared with T-Judge, OVD, and GAD. In the matched white-box setting, Qwen3-30B-A3B~\citep{yang2025qwen3technicalreport} serves as the open-weight teacher and is compared with SFT, the logit-based OPD baseline (hereafter LOPD), and ExOPD; LOPD and ExOPD use teacher logits, while ROPD receives teacher text only. Baseline details are in Appendix~\ref{app:baselines}.

\textbf{Data.} We use DAPO-Math-17K~\citep{yu2025dapo} for mathematics and RaR-Science/Medical-20K~\citep{gunjal2025rar} for science and medical tracks. All methods share the same training samples within each domain. SFT uses static teacher responses.
\textbf{Training.} All RL methods share the same GRPO recipe: learning rate $10^{-6}$, batch size 32, $n=8$ rollouts per prompt, and one epoch. ROPD uses $m=4$ teacher responses and $K \in [4, 12]$ rubric items. Checkpoints are selected using the validation split of the corresponding training dataset. Full settings are listed in Appendix~\ref{app:hyperparams}.

\textbf{Evaluation.} We evaluate on AIME24/25~\citep{aime2024,aime2025}, HMMT25~\citep{hmmt2025}, GPQA-Diamond~\citep{rein2023gpqa}, HealthBench~\citep{arora2025healthbench}, and IFEval~\citep{zhou2023instructionfollowingevaluation}. We sample $k=16$ responses per problem with temperature $1.0$, top-$p$ $0.95$, and a $32{,}768$-token cap. Teacher evaluation uses the same protocol; full details are provided in Appendix~\ref{app:evaluation_details}.

\subsection{Performance in Black-Box Scenarios}
\label{sec:blackbox-results}

\begin{table}[t]
\centering
\caption{Black-box distillation results (Pass@1\%). Bold/underline: best/second-best.}
\label{tab:main}
\footnotesize
\scoreinlinetrue
\setlength{\tabcolsep}{2pt}
\renewcommand{\arraystretch}{1.15}
\begin{tabular}{llllllll}
\toprule
& AIME24 & AIME25 & \makecell[l]{HMMT25\\(Feb.)} & \makecell[l]{HMMT25\\(Nov.)} & GPQA-D. & HealthBench & IFEval \\
\midrule
GPT-5.2-chat (teacher)  & 80.83 & 67.08 & 43.75 & 57.50 & 78.66 & 92.82 & 94.37 \\
\midrule
\multicolumn{8}{l}{\textbf{Non-Thinking}} \\
\midrule
Qwen3-4B (student)      & 24.17 & 20.83 & 10.42 & 7.08  & 35.66 & 83.32 & \underline{85.21} \\
T-Judge            & \secondup{62.50}{+38.3} & \secondup{56.64}{+35.8} & \scoreup{28.94}{+18.5} & \secondup{38.75}{+31.7} & \secondup{36.29}{+0.63} & \secondup{84.52}{+1.20} & \scoredown{84.40}{-0.81} \\
OVD~\citep{xiong2026ovd} & \scoreup{61.56}{+37.4} & \scoreup{55.71}{+34.9} & \secondup{29.11}{+18.7} & \scoreup{37.92}{+30.8} & \scoreup{35.74}{+0.08} & \scoreup{83.68}{+0.36} & \scoredown{84.23}{-0.98} \\
GAD~\citep{ye2026gad}   & \scoreup{27.52}{+3.35} & \scoreup{23.34}{+2.51} & \scoreup{12.84}{+2.42} & \scoreup{14.11}{+7.03} & \scoreup{36.02}{+0.36} & \scoreup{83.57}{+0.25} & \scoredown{85.12}{-0.09} \\
\rowcolor{BOPDRowBlue}
ROPD (ours)        & \bestup{65.02}{+40.9} & \bestup{58.75}{+37.9} & \bestup{31.69}{+21.3} & \bestup{41.67}{+34.6} & \bestup{36.50}{+0.84} & \bestup{84.92}{+1.60} & \bestup{85.28}{+0.07} \\
\bottomrule
\end{tabular}
\PrintScoreAnnotations
\renewcommand{\arraystretch}{1}
\end{table}

Table~\ref{tab:main} reports Pass@1 across seven benchmark configurations. ROPD ranks first in every configuration and obtains its largest gain on HMMT25 (Nov.), improving the base student from 7.08 to 41.67 (+34.6). The small IFEval gain indicates that this improvement does not reduce the evaluated instruction-following behavior; the corresponding thinking-mode results are reported in Table~\ref{tab:app_blackbox_thinking}.

\subsection{Matched White-Box Comparison}
\label{subsec:whitebox}
\label{subsec:efficiency}

\begin{table}[t]
\centering
\caption{Logit-based OPD results (Pass@1\%). Bold/underline: best/second-best.}
\label{tab:whitebox}
\footnotesize
\scoreinlinetrue
\setlength{\tabcolsep}{1.6pt}
\renewcommand{\arraystretch}{1.2}
\begin{tabular}{lclllll}
\toprule
& Access & AIME24 & AIME25 & \makecell[l]{HMMT25\\(Feb.)} & \makecell[l]{HMMT25\\(Nov.)} & Avg \\
\midrule
Qwen3-30B-A3B (teacher)            & --                   & 76.25 & 61.25 & 33.33 & 55.00 & 56.46 \\
\midrule
Qwen3-4B (student)                & --                   & 24.17 & 20.83 & 10.42 & 7.08 & 15.63 \\
SFT                               & text                 & \scoreup{26.69}{+2.52} & \scoreup{22.50}{+1.67} & \scoreup{11.62}{+1.20} & \scoreup{8.33}{+1.25} & \scoreup{17.29}{+1.66} \\
LOPD~\citep{agarwal2024gkd,lu2025onpolicydistillation}                         & logit              & \scoreup{47.92}{+23.8} & \scoreup{38.75}{+17.9} & \scoreup{20.42}{+10.0} & \scoreup{24.17}{+17.1} & \scoreup{32.82}{+17.2} \\
ExOPD~\citep{ExOPD}                             & logit              & \secondup{50.66}{+26.5} & \secondup{41.25}{+20.4} & \secondup{22.42}{+12.0} & \secondup{26.68}{+19.6} & \secondup{35.25}{+19.6} \\
\rowcolor{BOPDRowBlue}
ROPD (ours) & \textbf{text} & \bestup{63.33}{+39.2} & \bestup{55.93}{+35.1} & \bestup{25.40}{+15.0} & \bestup{38.80}{+31.7} & \bestup{45.87}{+30.2} \\
\bottomrule
\end{tabular}
\PrintScoreAnnotations
\renewcommand{\arraystretch}{1}
\end{table}


Table~\ref{tab:whitebox} reports the matched comparison with Qwen3-30B-A3B. Although ROPD uses teacher text only, it outperforms LOPD and ExOPD on all four mathematics benchmarks, closing 74.1\% of the student--teacher gap on average compared with 42.1\% for LOPD. Section~\ref{sec:reward_alignment} investigates why text-only rubric supervision can outperform dense logit access. ROPD is also more sample-efficient in this matched setting: it reaches LOPD's best AIME24 accuracy with 1.6k training samples versus 15.4k, and in 5.5h versus 34.4h, despite the added Rubricator and Verifier calls per step (Appendix~\ref{app:extra_figures} reports full curves, post-saturation behavior, and API cost).

\section{Analysis}
\label{sec:analysis}

Having established ROPD's empirical effectiveness, we now examine the properties of its reward signal. We begin with a concrete case illustrating how rubric-based rewards differ from scalar judges in distinguishing rollout quality (Section~\ref{sec:case_study}). We then quantify the alignment between reward signals and ground-truth correctness on a fixed rollout pool (Section~\ref{sec:reward_alignment}). Finally, we ablate the core design choices to assess the contribution of each reward component (Section~\ref{sec:ablation}).

\subsection{Case Study: Rubric \textit{vs.} Scalar Judge}
\label{sec:case_study}


Before turning to large-scale quantitative analysis, we use a single example to build intuition for how rubric-based and scalar rewards differ in discriminating rollout quality. Table~\ref{tab:case_study} presents a representative case regarding the parity-based contradiction: $n^3+3n^2+2n+1 \equiv 0 \pmod{2024}$. Since $n(n+1)(n+2)$ is inherently even, the expression remains odd, precluding any solution for the even modulus 2024.

We compare two student rollouts: Rollout A (partially correct, missing general parity proof) and Rollout C (fabricated derivation guessing $337$). The rubric separates them sharply ($0.77$ vs. $0.23$, $\Delta=0.54$), while the scalar judge barely distinguishes them ($0.70$ vs. $0.55$, $\Delta=0.15$).

This $3.6\times$ wider margin reflects a structural difference: scalar judges compress quality dimensions into a single value, allowing formatting to dilute logical failure. The rubric decouples evaluation dimensions (factorization, coherence, accuracy), preventing fabricated derivations from hiding behind structured prose.

The generality of this pattern is examined in Section~\ref{sec:reward_alignment} and the Appendix.

\begin{keyinsightbox}[title=Key Insight]
\textbf{Criterion-wise rubrics preserve correctness-relevant distinctions that scalar scores can compress.}
In this case, they separate a partially correct rollout from a fluent but fabricated one by a $3.6\times$ wider margin than scalar judging.
\end{keyinsightbox}




\subsection{Text-only Rubric Signals vs. Logit-based Supervision}
\label{sec:reward_alignment}



\begin{figure}[t]
\centering
\includegraphics[width=0.92\textwidth]{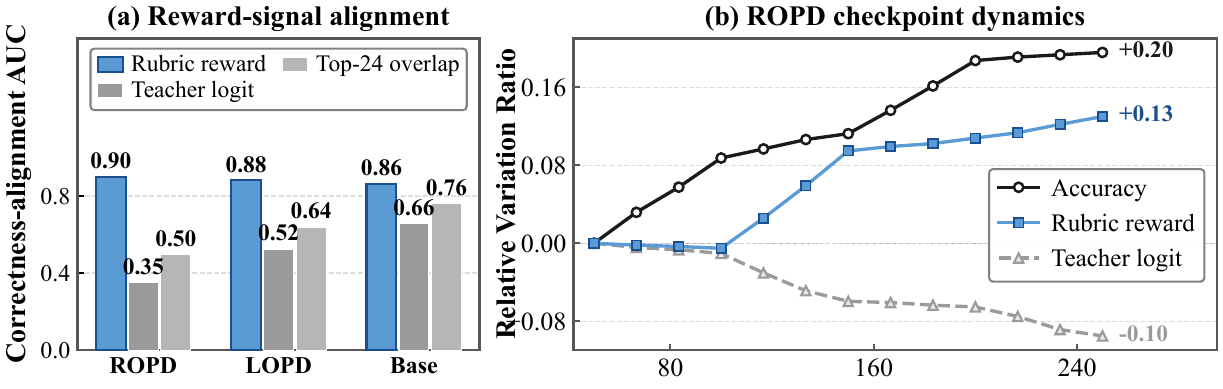}
\vspace{-2mm}
\caption{Reward signal alignment with correctness (AIME24). (a) Correctness-alignment AUC for rubric reward, teacher logit, and top-24 overlap. (b) Training trajectories: ROPD accuracy and rubric reward scale synchronously, while teacher logit diverges downward.}
\label{fig:signal_alignment}
\end{figure}

\begin{table}[t]
\centering
\caption{
  Case study: Multi-dimensional rubric evaluation on an AIME-style number theory problem. We present five rubrics alongside blind Verifier verdicts ($\checkmark$/$\times$) for two representative rollouts (A and C) selected from a group of eight. 
  Weights $w_k \in [1, 5]$ are dynamically assigned by the Rubricator.
}
\label{tab:case_study}
\footnotesize
\setlength{\tabcolsep}{1.5pt}
\begin{tabular}{l l p{4.8cm} c c c}
\toprule
ID & Category & Rubric & $w_k$ & Rollout~A & Rollout~C \\
\midrule
C1 & Task Completion    & Produces an explicit final answer.                                  & 5 & $\checkmark$ & $\checkmark$ \\
C2 & Observable Quality & Identifies the parity obstruction ($P(n)$ odd, 2024 even $\to$ no solution). & 5 & $\times$     & $\times$ \\
C3 & Observable Quality & Correctly factorizes $n^3+3n^2+2n$ into $n(n+1)(n+2)$.            & 4 & $\checkmark$ & $\times$ \\
C4 & General Reasoning  & Argument is logically coherent, each step follows from the last.    & 5 & $\checkmark$ & $\times$ \\
C5 & Observable Quality & No hallucinated numerical claims or guessed answers.                & 3 & $\checkmark$ & $\times$ \\
\midrule
\multicolumn{4}{l}{{Rubric Weighted Pass Rate}\,\,$\big(\sum_k w_k v_{i,k}\big/\sum_k w_k\big)$} & $17/22{=}0.77$ & $5/22{=}0.23$ \\
\cmidrule(lr){1-6}
\multicolumn{4}{l}{{Scalar Score}}                                                       & $0.70$         & $0.55$ \\
\bottomrule
\end{tabular}
\end{table}

Table~\ref{tab:whitebox} shows that text-only ROPD outperforms logit-based LOPD by +13.1 points on average. A natural question is whether this gap reflects differences in the quality of the underlying reward signal. We investigate this through a controlled analysis on a fixed pool of AIME24 rollouts, comparing three signals: (1)~rubric rewards, (2)~teacher log-likelihood, and (3)~top-24 token overlap, each evaluated against ground-truth correctness. Full protocol details are provided in Appendix~\ref{app:extra_figures}.

\begin{figure}[t]
\centering
\includegraphics[width=0.92\textwidth]{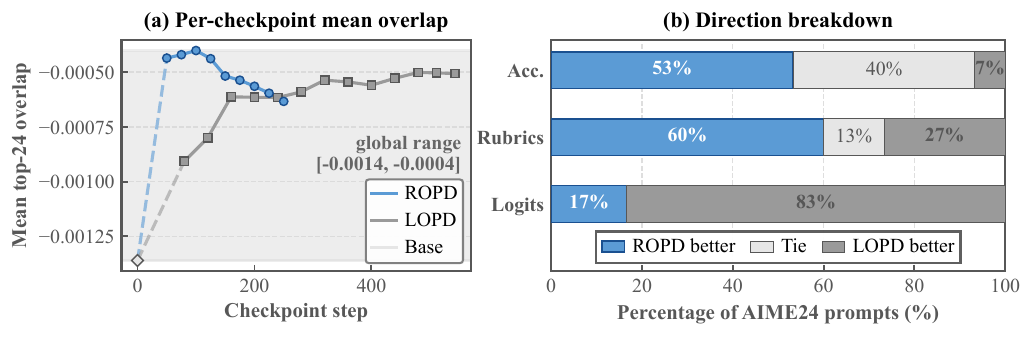}
\vspace{-2mm}
\caption{Evolution of stylistic mimicry and comparative performance. (a) Mimicry Trajectories: ROPD rapidly saturates stylistic alignment then pivots toward reasoning, while LOPD exhibits persistent mimicry. (b) Prompt-wise Comparison on AIME24: ROPD outperforms in reasoning accuracy and rubric satisfaction; LOPD's advantage is confined to logit mimicry.}
\label{fig:new_add}
\end{figure}

\textbf{Logit Likelihood is Poorly Aligned with Correctness.}
While LOPD treats teacher likelihood as a quality proxy, Figure~\ref{fig:signal_alignment}~(a) reveals that this signal is negatively associated with correctness: rubric rewards achieve an AUC of 0.90, whereas teacher log-likelihood yields an AUC of 0.35, below the 0.50 random baseline.
This pattern suggests that high teacher likelihood can favor fluent but logically flawed reasoning paths over correct but stylistically divergent ones. As shown in Figure~\ref{fig:new_add}~(b), ROPD produces more discriminative advantage signals across the majority of prompts. These observations are consistent with the rubric reward providing the optimizer with more correctness-aligned group-relative feedback than teacher log-likelihood.

\textbf{Early Convergence Followed by Divergence in Token-level Similarity.}
The training trajectories reveal a two-phase pattern in how ROPD relates to teacher token distributions. Figure~\ref{fig:new_add}~(a) shows that in the earliest stages, ROPD's token overlap increases even faster than LOPD's, suggesting that the rubric reward initially encourages alignment with the teacher's formatting and linguistic conventions.
However, as Figures~\ref{fig:new_add}~(a) and~\ref{fig:signal_alignment}~(b) show, a divergence follows: LOPD continues to optimize its logit-based objective throughout training, while ROPD's accuracy and rubric rewards increase as its token-level similarity to the teacher declines. One interpretation is that ROPD leverages teacher knowledge structurally rather than distributionally: once the student acquires the teacher's basic reasoning conventions, the rubric reward directs optimization toward correctness rather than token-level fidelity.

\textbf{Decoupled Supervision and Per-Criterion Progress.} The alignment evidence above concerns rollout-level reward signals. We next examine whether this alignment translates into targeted per-criterion gains during training.

To analyze criterion-level dynamics, we group criteria into three categories: \textit{Task Completion} (producing a required final answer), \textit{Observable Quality} (verifiable correctness evidence such as key steps, constraint identification, valid transformations, and absence of hallucinated claims), and \textit{General Reasoning} (logical consistency, self-checking, step-to-step connectivity). Category definitions and generation instructions are in Appendix~\ref{app:extra_figures}.

Table~\ref{tab:criterion_trajectory} reports pass rates across these categories. ROPD achieves superior gains ($\Delta$) in every dimension. By decomposing quality into independent milestones, the rubric enables separable feedback along each dimension, unlike LOPD's entangled logit signal. Table~\ref{tab:app_transition} shows a 15.9\% regression rate for LOPD's previously-passed cells, consistent with interference across dimensions.

Taken together, these results show that rubric rewards are more correctness-aligned than teacher log-likelihood and that ROPD's criterion-level progress is more stable. Which aspect of rubric construction contributes to these gains, however, remains an open question; Section~\ref{sec:ablation} examines it through controlled ablation of the reward signal's components.

\begin{keyinsightbox}[title=Key Insight]
\textbf{Under matched training, correctness-aligned rubric rewards stabilize criterion-level progress relative to logit-based supervision.}
On a fixed rollout pool, rubric rewards achieve an AUC of $0.90$, compared with $0.35$ for teacher log-likelihood; during training, ROPD improves more failed criteria and regresses fewer previously passed criteria than LOPD.
\end{keyinsightbox}


\vspace{-5pt}

\begin{table}[t]
\centering
\caption{Rubric-level pass rates of ROPD \textit{vs.} LOPD at early and final checkpoints on AIME24.}
\label{tab:criterion_trajectory}
\vspace{3pt}
\footnotesize
\setlength{\tabcolsep}{3pt}
\setlength{\extrarowheight}{1.5pt}
\renewcommand{\arraystretch}{1.08}
\begin{tabular}{lcccccc}
\toprule
& \multicolumn{3}{c}{ROPD} & \multicolumn{3}{c}{LOPD} \\
\cmidrule(r){2-4} \cmidrule(r){5-7}
Rubric Category & Early & Final & $\Delta$ & Early & Final & $\Delta$ \\
\midrule
Task Completion    & 54.0 & 67.6 & $+13.6$ & 48.0 & 53.3 & $+5.3$ \\
Observable Quality & 53.5 & 66.1 & $+12.6$ & 45.2 & 54.7 & $+9.5$ \\
General Reasoning  & 44.6 & 58.9 & $+14.3$ & 33.9 & 45.1 & $+11.2$ \\
\rowcolor{BOPDRowBlue}
\textbf{Overall}   & \textbf{52.5} & \textbf{65.6} & $\mathbf{+13.1}$ & \textbf{44.7} & \textbf{53.0} & $\mathbf{+8.3}$ \\
\bottomrule
\end{tabular}
\renewcommand{\arraystretch}{1.0}
\setlength{\extrarowheight}{0pt}
\PrintScoreAnnotations
\end{table}


\subsection{Ablation Study: Deconstructing the Reward Signal}
\label{sec:ablation}

We disentangle these contributions through a two-part ablation (Table~\ref{tab:reward_ablation}): \textit{rubric construction source ablation} varies what information the Rubricator receives, and \textit{leave-one-out component ablation} removes internal design choices from the full ROPD configuration.

\paragraph{Rubric construction source ablation.}

These variants share the same training infrastructure; only the Rubricator's input differs. Starting from the Generic baseline (60.79), adding teacher responses (Teacher-only 61.39) yields +0.60, while adding student rollouts (Student-only 62.25) yields +1.46. The larger effect from student conditioning suggests that anchoring criteria to the student's actual behavior produces more discriminative signals.

Actual ROPD (65.02) exceeds the additive baseline (62.85) by 2.17 points, suggesting synergistic interaction: the Rubricator requires both the teacher's quality principles and the student's error patterns. All partial-information variants (60.79--62.25) substantially underperform full ROPD (65.02).

\begin{table}[t]
\centering
\caption{Reward signal ablation on AIME24 (non-thinking, GPT-5.2-chat teacher). Pass@1 (\%) under (a) rubric construction sources and (b) leave-one-out design choices. T-Judge uses the same teacher as a holistic judge, controlling for judge strength; Teacher-only corresponds to RaR-style reference-grounded rubrics~\citep{gunjal2025rar}; Generic and Teacher-only rubrics are generated once before training.}
\label{tab:reward_ablation}
\small
\setlength{\tabcolsep}{12pt}
\renewcommand{\arraystretch}{1.3}
\begin{tabular}{llc}
\toprule
Variant & Configuration & AIME24 \\
\midrule
\multicolumn{3}{l}{\textit{(a) Rubric construction source}} \\
T-Judge       & teacher holistic score (no rubric)   & \scoreup{62.50}{+38.33} \\
Generic       & prompt only                          & \scoreup{60.79}{+36.62} \\
Teacher-only  & prompt + teacher responses           & \scoreup{61.39}{+37.22} \\
Student-only  & prompt + student rollouts            & \scoreup{62.25}{+38.08} \\
\midrule
\multicolumn{3}{l}{\textit{(b) Leave-one-out internal design}} \\
Qwen3-4B (base) & (no RL training)                   & 24.17 \\
w/o multi-teacher & single teacher answer ($m=1$)    & \scoreup{47.08}{+22.91} \\
w/o sharing       & per-student rubrics              & \scoreup{61.25}{+37.08} \\
w/o blind scoring & verifier sees response source    & \scoreup{61.75}{+37.58} \\
\midrule
\rowcolor{BOPDRowBlue}
\textbf{ROPD (full)} & \textbf{prompt + teacher + student, $m=4$} & \bestup{65.02}{+40.85} \\
\bottomrule
\end{tabular}
\renewcommand{\arraystretch}{1.0}
\PrintScoreAnnotations
\end{table}

\paragraph{Leave-one-out component ablation.}
Given the rubric construction source, ROPD's reward signal further depends on three internal design choices: multi-teacher seeding, cross-rollout rubric sharing, and blind verification. Table~\ref{tab:reward_ablation}b presents a leave-one-out ablation.
\textbf{Multi-teacher coverage is the primary driver.} Transitioning from $m=4$ to $m=1$ results in a 17.9-point drop; diverse strategies help induce path-independent criteria.
\textbf{Sharing aggregates cross-rollout contrast.} A single shared rubric per prompt yields +3.77 points, surfacing systematic gaps that individual rubrics miss.
\textbf{Blind scoring prevents identity bias.} Revealing response identities costs 3.27 points; the Verifier evaluates rollouts without knowing their source.

\begin{keyinsightbox}[title=Key Insight]
\textbf{Teacher--student contrastive rubric construction is central to reward quality.}
Using both teacher responses and current student rollouts to construct rubrics outperforms all partial-information variants; multi-teacher coverage is the dominant component, while shared rubrics and blind verification provide complementary gains.
\end{keyinsightbox}

\section{Related Work}
\label{sec:related}


\vspace{-2pt}


\paragraph{On-policy Distillation.}
On-policy distillation (OPD) trains a student on feedback computed over its own rollouts, mitigating exposure bias and improving sample efficiency~\citep{gu2024minillm,agarwal2024gkd,lu2025onpolicydistillation,song2026surveyopd}. Recent work strengthens OPD through objective design and reward extrapolation~\citep{jin2026entropy,ExOPD}, training efficiency and signal calibration~\citep{zhang2026fast,wu2026lightning,xu2026paced,xu2026tip,zheng2026scope}, cross-tokenizer distillation~\citep{zhang2025dual}, and empirical analyses of failure modes~\citep{li2026rethinkingopd,fu2026revisiting}. These methods require dense teacher logits, which proprietary teachers do not expose.

\vspace{-3pt}

\paragraph{Black-box Distillation.}
When teachers expose only text, black-box distillation substitutes response-level signals for logits. SeqKD on teacher demonstrations~\citep{kim2016sequence} and preference optimization on fixed pairs~\citep{singh2025orpodistill} remain offline. A narrower set preserves on-policy updates on current student rollouts: GAD (a co-evolving discriminator)~\citep{ye2026gad}, OVD (verbal trajectory scores)~\citep{xiong2026ovd}, and RL-based KD (a judge's continuous reward)~\citep{shen2026rljudgekd}. These methods show that text-only teachers can supervise current rollouts, but collapse each response into a single preference or score.

\vspace{-3pt}


\paragraph{Rubric-based Reinforcement Learning.} Rubric-based reinforcement learning decomposes response quality into explicit criteria, extending RLVR beyond binary-correctness domains~\citep{shao2024deepseekmath}. RaR~\citep{gunjal2025rar}, OpenRubrics~\citep{zhou2026openrubrics}, Rubicon~\citep{huang2025rubicon}, RLER~\citep{shao2025rler}, and SibylSense~\citep{xu2026sibylsense} formalize, scale, and evolve these criteria, building rubrics from reference answers, task descriptions, or accumulated evidence. In these methods, rubrics define the target quality of a task. ROPD shares their criterion-level rewards and uses rubrics for a different purpose, transferring a teacher's competence: criteria are induced from the contrast between multiple teacher responses and the current student rollout group, making rubrics a text-only, teacher-grounded, student-conditioned distillation interface.

\section{Conclusion}

We studied whether rubric-based supervision can serve as a text-only feedback interface for black-box on-policy distillation, using ROPD as a concrete instantiation. Under matched settings, ROPD outperforms logit-based OPD on all four math benchmarks and reaches the best AIME24 accuracy of standard logit-based OPD with 9.6$\times$ fewer training samples. We trace this result to correctness-aligned rubric rewards, stable criterion-level progress, and teacher--student contrastive rubric construction. These findings establish the interface in the evaluated settings; extending it to open-ended settings, where criteria are harder to verify, and to broader student families remains future work.

\subsection*{AI use statement}

We used LLM-based tools to aid and polish the writing of this paper,
including editing draft text for clarity and flow, and proofreading.
All technical content, experiments, and conclusions are the work of
the authors. We did not use LLMs for data generation, methodology or
experiment design, method implementation, or analysis of results.

We have reviewed all AI-assisted work. We take responsibility for the
final content of this work.

\subsection*{Reproducibility statement}

We provide a complete specification of the ROPD training procedure and
evaluation protocol. Appendix~\ref{app:algorithm} presents the full
pseudocode of the training loop, including the rubric induction, verification,
and GRPO update stages. Appendix~\ref{app:hyperparams} lists all training
hyperparameters, and Appendix~\ref{app:prompts} provides the exact prompt
templates used to elicit rubrics and verification judgements, so the pipeline
can be reimplemented with any text-only teacher. Section~\ref{sec:setup}
describes the datasets, benchmarks, and baselines;
Appendix~\ref{app:evaluation_details} documents the decoding and evaluation
settings, and Appendix~\ref{app:qualitative} includes worked rubric examples
and case studies that illustrate the reward construction. All models and
datasets used in this paper are publicly available.

\bibliographystyle{plainnat}
\bibliography{reference}

@article{ExOPD,
  author       = {Wenkai Yang and
                  Weijie Liu and
                  Ruobing Xie and
                  Kai Yang and
                  Saiyong Yang and
                  Yankai Lin},
  title        = {Learning beyond Teacher: Generalized On-Policy Distillation with Reward
                  Extrapolation},
  journal      = {CoRR},
  volume       = {abs/2602.12125},
  year         = {2026}
}

@misc{openai2025gpt52,
  title        = {Introducing GPT-5.2},
  author       = {{OpenAI}},
  year         = {2025},
  howpublished = {\url{https://openai.com/index/introducing-gpt-5-2/}},
  note         = {Accessed: 2026-05-06}
}

@article{gemma3,
  title   = {Gemma 3 Technical Report},
  author  = {{Gemma Team, Google DeepMind}},
  journal = {arXiv preprint arXiv:2503.19786},
  year    = {2025},
  url     = {https://arxiv.org/abs/2503.19786}
}

@misc{lu2025onpolicydistillation,
  title        = {On-policy distillation},
  author       = {Lu, Kevin and Thinking Machines Lab},
  year         = {2025},
  howpublished = {\emph{Thinking Machines Lab: Connectionism}},
  doi          = {10.64434/tml.20251026},
  url          = {https://thinkingmachines.ai/blog/on-policy-distillation}
}

@article{hinton2015distilling,
  title = {Distilling the Knowledge in a Neural Network},
  author = {Hinton, Geoffrey and Vinyals, Oriol and Dean, Jeff},
  journal = {arXiv preprint arXiv:1503.02531},
  year = {2015},
  url = {https://arxiv.org/abs/1503.02531}
}

@inproceedings{kim2016sequence,
  title = {Sequence-Level Knowledge Distillation},
  author = {Kim, Yoon and Rush, Alexander M.},
  booktitle = {Proceedings of the 2016 Conference on Empirical Methods in Natural Language Processing},
  year = {2016},
  pages = {1317--1327},
  url = {https://aclanthology.org/D16-1139/}
}

@inproceedings{gu2024minillm,
  title = {MiniLLM: Knowledge Distillation of Large Language Models},
  author = {Gu, Yuxian and Dong, Li and Wei, Furu and Huang, Minlie},
  booktitle = {International Conference on Learning Representations},
  year = {2024},
  url = {https://arxiv.org/abs/2306.08543}
}

@inproceedings{agarwal2024gkd,
  title = {On-Policy Distillation of Language Models: Learning from Self-Generated Mistakes},
  author = {Agarwal, Rishabh and Vieillard, Nino and Stanczyk, Piotr and Ramos, Sabela and Geist, Matthieu and Bachem, Olivier},
  booktitle = {International Conference on Learning Representations},
  year = {2024},
  url = {https://arxiv.org/abs/2306.13649}
}

@article{yang2025qwen3technicalreport,
  title = {Qwen3 Technical Report},
  author = {Yang, An and Li, Anfeng and Yang, Baosong and Zhang, Beichen and Hui, Binyuan and Zheng, Bo and Yu, Bowen and Gao, Chang and Huang, Chengen and Lv, Chenxu and Zheng, Chujie and Liu, Dayiheng and others},
  journal = {arXiv preprint arXiv:2505.09388},
  year = {2025},
  url = {https://arxiv.org/abs/2505.09388}
}

@article{li2026rethinkingopd,
  title = {Rethinking On-Policy Distillation of Large Language Models: Phenomenology, Mechanism, and Recipe},
  author = {Li, Yaxuan and Zuo, Yuxin and He, Bingxiang and Zhang, Jinqian and Xiao, Chaojun and Qian, Cheng and Yu, Tianyu and Gao, Huan-ang and Yang, Wenkai and Liu, Zhiyuan and Ding, Ning},
  journal = {arXiv preprint arXiv:2604.13016},
  year = {2026},
  url = {https://arxiv.org/abs/2604.13016}
}

@article{song2026surveyopd,
  title = {A Survey of On-Policy Distillation for Large Language Models},
  author = {Song, Mingyang and Zheng, Mao},
  journal = {arXiv preprint arXiv:2604.00626},
  year = {2026},
  url = {https://arxiv.org/abs/2604.00626}
}

@article{ye2026gad,
  title = {Black-Box On-Policy Distillation of Large Language Models},
  author = {Ye, Tianzhu and Dong, Li and Chi, Zewen and Wu, Xun and Huang, Shaohan and Wei, Furu},
  journal = {arXiv preprint arXiv:2511.10643},
  year = {2026},
  url = {https://arxiv.org/abs/2511.10643}
}

@article{singh2025orpodistill,
  title = {ORPO-Distill: Mixed-Policy Preference Optimization for Cross-Architecture LLM Distillation},
  author = {Singh, Aasheesh and Vaddina, Vishal and Birru, Dagnachew},
  journal = {arXiv preprint arXiv:2509.25100},
  year = {2025},
  url = {https://arxiv.org/abs/2509.25100}
}

@article{xiong2026ovd,
  title = {OVD: On-policy Verbal Distillation},
  author = {Xiong, Jing and Shen, Hui and Gong, Shansan and Cheng, Yuxin and Shen, Jianghan and Tao, Chaofan and Tan, Haochen and Bai, Haoli and Shang, Lifeng and Wong, Ngai},
  journal = {arXiv preprint arXiv:2601.21968},
  year = {2026},
  url = {https://arxiv.org/abs/2601.21968}
}

@article{shen2026rljudgekd,
  title = {Reinforcement Learning-based Knowledge Distillation with LLM-as-a-Judge},
  author = {Shen, Yiyang and Tu, Lifu and Wang, Weiran},
  journal = {arXiv preprint arXiv:2604.02621},
  year = {2026},
  url = {https://arxiv.org/abs/2604.02621}
}

@article{shao2024deepseekmath,
  title = {DeepSeekMath: Pushing the Limits of Mathematical Reasoning in Open Language Models},
  author = {Shao, Zhihong and Wang, Peiyi and Zhu, Qihao and Xu, Runxin and Song, Junxiao and Bi, Xiao and Zhang, Haowei and Zhang, Mingchuan and Li, Y. K. and Wu, Y. and Guo, Daya},
  journal = {arXiv preprint arXiv:2402.03300},
  year = {2024},
  url = {https://arxiv.org/abs/2402.03300}
}

@article{li2026rubricarm,
  title = {Alternating Reinforcement Learning for Rubric-Based Reward Modeling in Non-Verifiable LLM Post-Training},
  author = {Xu, Ran and Liu, Tianci and Dong, Zihan and Yu, Tony and Hong, Ilgee and Yang, Carl and Zhang, Linjun and Zhao, Tao and Wang, Haoyu},
  journal = {arXiv preprint arXiv:2602.01511},
  year = {2026},
  url = {https://arxiv.org/abs/2602.01511}
}

@article{zhou2026openrubrics,
  title={Openrubrics: Towards scalable synthetic rubric generation for reward modeling and llm alignment},
  author={Liu, Tianci and Xu, Ran and Yu, Tony and Hong, Ilgee and Yang, Carl and Zhao, Tuo and Wang, Haoyu},
  journal={arXiv preprint arXiv:2510.07743},
  year={2025}
}

@article{gunjal2025rar,
  title = {Rubrics as Rewards: Reinforcement Learning Beyond Verifiable Domains},
  author = {Gunjal, Anisha and Wang, Anthony and Lau, Elaine and Nath, Vaskar and He, Yunzhong and Liu, Bing and Hendryx, Sean},
  journal = {arXiv preprint arXiv:2507.17746},
  year = {2025},
  url = {https://arxiv.org/abs/2507.17746}
}

@article{huang2025rubicon,
  title = {Reinforcement Learning with Rubric Anchors},
  author = {Huang, Zenan and Zhuang, Yihong and Lu, Guoshan and Qin, Zeyu and Xu, Haokai and Zhao, Tianyu and Peng, Ru and Hu, Jiaqi and Shen, Zhanming and Hu, Xiaomeng and Gu, Xijun and Tu, Peiyi and Liu, Jiaxin and Chen, Wenyu and Fu, Yuzhuo and Fan, Zhiting and Gu, Yanmei and Wang, Yuanyuan and Yang, Zhengkai and Li, Jianguo and Zhao, Junbo},
  journal = {arXiv preprint arXiv:2508.12790},
  year = {2025},
  url = {https://arxiv.org/abs/2508.12790}
}

@article{shao2025rler,
  title = {DR Tulu: Reinforcement Learning with Evolving Rubrics for Deep Research},
  author = {Shao, Rulin and Asai, Akari and Shen, Shannon Zejiang and Ivison, Hamish and Kishore, Varsha and Zhuo, Jingming and Zhao, Xinran and Park, Molly and Finlayson, Samuel G. and Sontag, David and Murray, Tyler and Min, Sewon and Dasigi, Pradeep and Soldaini, Luca and Brahman, Faeze and Yih, Wen-tau and Wu, Tongshuang and Zettlemoyer, Luke and Kim, Yoon and Hajishirzi, Hannaneh and Koh, Pang Wei},
  journal = {arXiv preprint arXiv:2511.19399},
  year = {2025},
  url = {https://arxiv.org/abs/2511.19399}
}

@article{xu2026sibylsense,
  title = {SibylSense: Adaptive Rubric Learning via Memory Tuning and Adversarial Probing},
  author = {Xu, Yifei and Potje, Guilherme and Shandilya, Shivam and Yuan, Tiancheng and de Oliveira Nunes, Leonardo and Agarwal, Rakshanda and Asgari, Saeid and Atkinson, Adam and K{\i}c{\i}man, Emre and Lu, Songwu and Chandra, Ranveer and Chakraborty, Tusher},
  journal = {arXiv preprint arXiv:2602.20751},
  year = {2026},
  url = {https://arxiv.org/abs/2602.20751}
}

@article{yu2025dapo,
  title = {DAPO: An Open-Source LLM Reinforcement Learning System at Scale},
  author = {Yu, Qiying and Zhang, Zheng and Zhu, Ruofei and Yuan, Yufeng and Zuo, Xiaochen and Yue, Yu and Dai, Weinan and Fan, Tiantian and Liu, Gaohong and Liu, Lingjun and Liu, Xin and Lin, Haibin and Lin, Zhiqi and Ma, Bole and Sheng, Guangming and Tong, Yuxuan and Zhang, Chi and Zhang, Mofan and Zhang, Wang and Zhu, Hang and Zhu, Jinhua and Chen, Jiaze and Chen, Jiangjie and Wang, Chengyi and Yu, Hongli and Song, Yuxuan and Wei, Xiangpeng and Zhou, Hao and Liu, Jingjing and Ma, Wei-Ying and Zhang, Ya-Qin and Yan, Lin and Qiao, Mu and Wu, Yonghui and Wang, Mingxuan},
  journal = {arXiv preprint arXiv:2503.14476},
  year = {2025},
}

@article{arora2025healthbench,
  title = {HealthBench: Evaluating Large Language Models Towards Improved Human Health},
  author = {Arora, Rahul K. and Wei, Jason and Hicks, Rebecca Soskin and Bowman, Preston and Qui{\~n}onero-Candela, Joaquin},
  journal = {arXiv preprint arXiv:2505.08775},
  year = {2025},
}

@article{rein2023gpqa,
  title  = {GPQA: A Graduate-Level Google-Proof Q\&A Benchmark},
  author = {Rein, David and Hou, Betty Li and Stickland, Asa Cooper and Petty, Jackson and Pang, Richard Yuanzhe and Dirani, Julien and Michael, Julian and Bowman, Samuel R.},
  journal= {arXiv preprint arXiv:2311.12022},
  year   = {2023},
}

@article{zhou2023instructionfollowingevaluation,
  title  = {Instruction-Following Evaluation for Large Language Models},
  author = {Zhou, Jeffrey and Lu, Tianjian and Mishra, Swaroop and Brahma, Siddhartha and Basu, Sujoy and Luan, Yi and Zhou, Denny and Hou, Le},
  journal= {arXiv preprint arXiv:2311.07911},
  year   = {2023},
}

@misc{aime2024,
  title  = {AIME 2024: American Invitational Mathematics Examination},
  author = {{MAA}},
  year   = {2024},
  url    = {https://artofproblemsolving.com/wiki/index.php/2024_AIME_I},
}

@misc{aime2025,
  title  = {AIME 2025: American Invitational Mathematics Examination},
  author = {{MAA}},
  year   = {2025},
  url    = {https://artofproblemsolving.com/wiki/index.php/2025_AIME_I},
}

@misc{hmmt2025,
  title  = {HMMT 2025: Harvard-MIT Mathematics Tournament},
  author = {{HMMT}},
  year   = {2025},
}

@article{xiao2026mimo,
  title={Mimo-v2-flash technical report},
  author={Xiao, Bangjun and Xia, Bingquan and Yang, Bo and Gao, Bofei and Shen, Bowen and Zhang, Chen and He, Chenhong and Lou, Chiheng and Luo, Fuli and Wang, Gang and others},
  journal={arXiv preprint arXiv:2601.02780},
  year={2026}
}

@article{deepseekai2026deepseekv4,
  title = {{DeepSeek-V4}: Towards Highly Efficient Million-Token Context Intelligence},
  author = {{DeepSeek-AI}},
  journal = {arXiv preprint arXiv:2606.19348},
  year = {2026},
  url = {https://arxiv.org/abs/2606.19348}
}

@article{fu2026revisiting,
  title={Revisiting On-Policy Distillation: Empirical Failure Modes and Simple Fixes},
  author={Fu, Yuqian and Huang, Haohuan and Jiang, Kaiwen and Liu, Jiacai and Jiang, Zhuo and Zhu, Yuanheng and Zhao, Dongbin},
  journal={arXiv preprint arXiv:2603.25562},
  year={2026}
}

@article{xu2026tip,
  title={TIP: Token Importance in On-Policy Distillation},
  author={Xu, Yuanda and Sang, Hejian and Zhou, Zhengze and He, Ran and Wang, Zhipeng and Geramifard, Alborz},
  journal={arXiv preprint arXiv:2604.14084},
  year={2026}
}

@article{zheng2026scope,
  title={SCOPE: Signal-Calibrated On-Policy Distillation Enhancement with Dual-Path Adaptive Weighting},
  author={Zheng, Binbin and Ma, Xing and Liang, Yiheng and Ruan, Jingqing and Fu, Xiaoliang and Lin, Kepeng and Zhu, Benchang and Zeng, Ke and Cai, Xunliang},
  journal={arXiv preprint arXiv:2604.10688},
  year={2026}
}

@article{jin2026entropy,
  title={Entropy-Aware On-Policy Distillation of Language Models},
  author={Jin, Woogyeol and Min, Taywon and Yang, Yongjin and Kadhe, Swanand Ravindra and Zhou, Yi and Wei, Dennis and Baracaldo, Nathalie and Lee, Kimin},
  journal={arXiv preprint arXiv:2603.07079},
  year={2026}
}

@article{zhang2026fast,
  title={Fast and Effective On-policy Distillation from Reasoning Prefixes},
  author={Zhang, Dongxu and Yang, Zhichao and Janghorbani, Sepehr and Han, Jun and Ressler II, Andrew and Qian, Qian and Lyng, Gregory D and Batra, Sanjit Singh and Tillman, Robert E},
  journal={arXiv preprint arXiv:2602.15260},
  year={2026}
}

@article{xu2026paced,
  title={PACED: Distillation and On-Policy Self-Distillation at the Frontier of Student Competence},
  author={Xu, Yuanda and Sang, Hejian and Zhou, Zhengze and He, Ran and Wang, Zhipeng},
  journal={arXiv preprint arXiv:2603.11178},
  year={2026}
}

@article{wu2026lightning,
  title={Lightning OPD: Efficient Post-Training for Large Reasoning Models with Offline On-Policy Distillation},
  author={Wu, Yecheng and Han, Song and Cai, Hai},
  journal={arXiv preprint arXiv:2604.13010},
  year={2026}
}

@article{zhang2025dual,
  title={A dual-space framework for general knowledge distillation of large language models},
  author={Zhang, Xue and Zhang, Songming and Liang, Yunlong and Meng, Fandong and Chen, Yufeng and Xu, Jinan and Zhou, Jie},
  journal={arXiv preprint arXiv:2504.11426},
  year={2025}
}

\newpage
\appendix
\section*{Appendix}
\addcontentsline{toc}{section}{Appendix}

\noindent\textbf{Appendix Overview}

\noindent \S\ref{app:broader_impact}~~Broader Impacts \dotfill \pageref{app:broader_impact}

\noindent \S\ref{app:related_full}~~Related Work (Complete Version) \dotfill \pageref{app:related_full}

\noindent \S\ref{app:qualitative}~~Qualitative Analysis and Case Studies \dotfill \pageref{app:qualitative}

\noindent \S\ref{app:hyperparams}~~Hyperparameters and Training Configuration \dotfill \pageref{app:hyperparams}

\noindent \S\ref{app:prompts}~~Prompt Templates \dotfill \pageref{app:prompts}

\noindent \S\ref{app:extra_figures}~~Additional Figures and Analysis \dotfill \pageref{app:extra_figures}

\noindent \S\ref{app:cross_arch}~~Cross-Architecture Generalization \dotfill \pageref{app:cross_arch}

\noindent \S\ref{app:algorithm}~~Algorithm Pseudocode and Method Details \dotfill \pageref{app:algorithm}

\vspace{8pt}

\newpage
\section{Broader Impacts}
\label{app:broader_impact}

ROPD studies a potentially more information-efficient interface for LLM alignment by substituting opaque token-level mimicry with structured, human-readable semantic rubrics. In the matched comparison reported in Section~\ref{subsec:efficiency}, this interface reaches the LOPD reference performance with approximately 9.6$\times$ fewer training samples, while API and serving costs remain practical considerations. By grounding the distillation process in explicit criteria rather than implicit probability distributions, ROPD offers enhanced transparency; developers can audit and intervene in the specific principles being transferred. However, the reliance on automated rubricators may inadvertently propagate or amplify biases inherent in the teacher's evaluative logic. If a teacher's rubrics contain flawed or prejudiced heuristics, the student may systematically internalize these errors under the guise of logical rules. Therefore, the deployment of rubric-based distillation should be accompanied by rigorous fairness audits of the induced criteria and the development of safety-aligned verifiers, especially in high-stakes sectors like medicine or law. We view ROPD as a starting point for studying text-only distillation, while emphasizing that the transition from blind imitation to principled alignment must be guided by ethical accountability and robust safety monitoring.


\newpage
\section{Related Work (Complete Version)}
\label{app:related_full}

This section provides the complete Related Work discussion with full context and citations.
A condensed overview appears in Section~\ref{sec:related} of the main text.

\paragraph{Knowledge distillation and on-policy distillation.}
Knowledge distillation (KD) transfers the behavior of a large teacher model into a smaller student, and is widely used to adapt or compress language models. Classical KD matches teacher soft targets on a fixed data distribution~\citep{hinton2015distilling}, and Sequence-Level Knowledge Distillation (SeqKD) extends this to generation by substituting teacher-decoded sequences for label-level targets~\citep{kim2016sequence}. Both are offline and suffer from exposure bias: training follows teacher-forced trajectories, while inference exposes the student to its own prefixes and errors, creating a mismatch between the distributions seen at training and test time. On-policy distillation (OPD) addresses this by training on student-generated sequences: MiniLLM optimizes reverse-KL on sampled responses~\citep{gu2024minillm}, Generalized Knowledge Distillation (GKD) learns from self-generated mistakes with teacher feedback~\citep{agarwal2024gkd}, and recent work scales this recipe to reasoning post-training~\citep{yang2025qwen3technicalreport,li2026rethinkingopd}. Despite this progress, these methods share a common assumption: they require token-level teacher information such as logits, which is unavailable for proprietary teachers and difficult to align across different architectures or vocabularies. ROPD studies the complementary black-box regime where the teacher exposes only text responses, enabling on-policy distillation when token-level supervision is infeasible.

\paragraph{Black-box On-policy Distillation.}
Recent black-box distillation methods answer this question with different forms of response-level supervision: ORPO-Distill constructs mixed-policy preference pairs from teacher and student reasoning traces~\citep{singh2025orpodistill}; GAD trains a discriminator to distinguish teacher from student responses and uses its score as a co-evolving reward~\citep{ye2026gad}; On-policy Verbal Distillation (OVD) asks the teacher for discrete verbal trajectory scores, avoiding token alignment and reducing memory cost~\citep{xiong2026ovd}; and RL-based KD with LLM-as-a-Judge trains from scalar evaluator rewards over unlabeled data~\citep{shen2026rljudgekd}. These methods demonstrate that output-only teachers can supervise student rollouts, but their signals remain largely implicit or holistic: preferences compare whole traces, discriminators hide the criteria behind a learned score, and verbal or judge rewards summarize a response into a single value. ROPD instead makes the distillation interface explicit by deriving shared rubrics from multiple teacher answers and current student rollouts, verifying each rollout against these criteria, and using the resulting weighted pass rates as on-policy rewards.

\paragraph{Rubric-based Reinforcement Learning.}
Reinforcement learning with verifiable rewards (RLVR) has driven strong gains in math and code~\citep{shao2024deepseekmath}, but its reliance on binary correctness limits it to domains with deterministic ground truth. Rubrics address this by decomposing response quality into structured, multi-dimensional criteria, extending RL to open-ended tasks. Rubrics-as-Rewards (RaR) formalized instance-specific rubrics as on-policy RL rewards, showing RLVR to be a special case of rubric-based RL~\citep{gunjal2025rar}. On rubric generation, OpenRubrics scales synthesis via contrastive prompting~\citep{zhou2026openrubrics}. On training dynamics, Rubicon identifies a seesaw effect between conflicting rubric types---improving one dimension can degrade another---and proposes multi-stage training to stabilize learning~\citep{huang2025rubicon}. Recognizing that static rubrics fail to capture emergent behaviors, Reinforcement Learning with Evolving Rubrics (RLER) and SibylSense introduce evolving rubrics that co-adapt with the policy: RLER grounds them on retrieved search evidence~\citep{shao2025rler}, while SibylSense pursues adversarial memory tuning~\citep{xu2026sibylsense}. A common assumption underlies these methods: rubrics function as evaluation instruments---they measure response quality against criteria sourced from benchmarks, reference answers, or self-generated preferences---but they are not designed to transfer knowledge from a stronger model to a weaker one. ROPD instead induces rubrics from the contrast between multi-teacher answers and on-policy student rollouts, converting them via a verifier into weighted pass-rate rewards for Group Relative Policy Optimization (GRPO). This repositions rubrics as a distillation interface---the resulting reward is simultaneously teacher-grounded and rollout-conditioned.

\newpage
\section{Qualitative Analysis and Case Studies}
\label{app:qualitative}

\paragraph{Case study: Rubric disagreement reveals teacher bias.}
When multiple teacher answers disagree on a rubric criterion, the Rubricator surfaces
this ambiguity explicitly (\textit{e.g.}, ``Criterion 7: Uses proof by induction -- 2/4 teachers
support, 2/4 use direct computation''). This prevents the student from overfitting to
one teacher's style.

\paragraph{Case study: Failure mode -- rubric exploitation.}
In rare cases ($<2\%$ of rollouts), the student learns to produce responses that score
highly on rubrics without being substantively correct (\textit{e.g.}, formatting tricks, keyword
stuffing). We observe this mainly in early training, and it diminishes as training proceeds:
each rubric contains at least one high-weight criterion that checks final-answer correctness
(Appendix~\ref{app:prompts}), which bounds the reward attainable by responses that satisfy
only surface-level criteria.

\paragraph{Rubric item examples.}
Table~\ref{tab:app_rubric_examples} shows representative rubric items generated by the
Rubricator for different prompt types.

\begin{table}[h]
\centering
\caption{\textbf{Representative rubric items generated by ROPD's Rubricator.}
$K=12$ items are generated per instance; we show 4 examples per domain.}
\label{tab:app_rubric_examples}
\small
\begin{tabular}{l p{7cm}}
\toprule
Domain & Example Rubric Items \\
\midrule
\multirow{4}{*}{Math (AIME)}
& ``The solution defines all variables before computation'' \\
& ``Intermediate steps are explicitly justified with theorems or algebraic rules'' \\
& ``The final answer is boxed and matches the required format'' \\
& ``No arithmetic errors in the numerical computation chain'' \\
\midrule
\multirow{4}{*}{Science (GPQA)}
& ``The answer identifies the relevant physical/chemical principle'' \\
& ``Quantitative reasoning includes correct unit conversions'' \\
& ``Alternative hypotheses are considered and ruled out'' \\
& ``The conclusion explicitly addresses the question asked'' \\
\midrule
\multirow{4}{*}{Medicine (HealthBench)}
& ``Diagnosis is supported by specific findings from the case description'' \\
& ``Differential diagnosis lists at least 2 alternative conditions'' \\
& ``Treatment recommendation follows guideline-concordant reasoning'' \\
& ``Referral or follow-up plan is specified when appropriate'' \\
\bottomrule
\end{tabular}
\end{table}

\newpage
\section{Hyperparameters and Training Configuration}
\label{app:hyperparams}

\paragraph{Complete hyperparameter specification.}
Table~\ref{tab:app_hyperparams} lists all hyperparameters used in ROPD experiments.

\begin{table}[h]
\centering
\caption{\textbf{Complete hyperparameter configuration.} For the matched white-box comparison (Table~\ref{tab:whitebox}), Qwen3-30B-A3B replaces GPT-5.2-chat-latest as the teacher, Rubricator, and Verifier on the math track; all other settings are unchanged.}
\label{tab:app_hyperparams}
\small
\begin{tabular}{l c@{\hskip 4pt}c@{\hskip 4pt}c}
\toprule
Hyperparameter & Math Track & Science Track & Medical Track \\
\midrule
\multicolumn{4}{c}{\textit{Model}} \\
Student model       & Qwen3-4B & Qwen3-4B & Qwen3-4B \\
Teacher model       & GPT-5.2-chat-latest & GPT-5.2-chat-latest & GPT-5.2-chat-latest \\
Rubricator model    & GPT-5.2-chat-latest & GPT-5.2-chat-latest & GPT-5.2-chat-latest \\
Verifier model      & GPT-5.2-chat-latest & GPT-5.2-chat-latest & GPT-5.2-chat-latest \\
\midrule
\multicolumn{4}{c}{\textit{Training}} \\
Training dataset    & DAPO-Math-17K & RaR-Science-20k & RaR-Medical-20k \\
Learning rate       & $1\times10^{-6}$ & $1\times10^{-6}$ & $1\times10^{-6}$ \\
LR scheduler        & Cosine & Cosine & Cosine \\
Warmup steps        & 100 & 100 & 100 \\
Batch size          & 32 & 32 & 32 \\
GRPO group size $n$ & 8 & 8 & 8 \\
Max training steps  & 531 & 625 & 625 \\
Precision           & bf16 & bf16 & bf16 \\
Optimizer           & AdamW & AdamW & AdamW \\
AdamW ($\beta_1$, $\beta_2$) & (0.9, 0.95) & (0.9, 0.95) & (0.9, 0.95) \\
Weight decay        & 0.1 & 0.1 & 0.1 \\
Gradient clipping   & 1.0 & 1.0 & 1.0 \\
\midrule
\multicolumn{4}{c}{\textit{ROPD Specific}} \\
Teacher answers $m$ & 4 & 4 & 4 \\
Rubric items $K$    & 4--12 & 4--12 & 4--12 \\
Rubricator temperature & 0.7 & 0.7 & 0.7 \\
Verifier temperature    & 0.0 & 0.0 & 0.0 \\
\midrule
\multicolumn{4}{c}{\textit{Training Rollout Decoding}} \\
Max rollout tokens & 8192 & 8192 & 8192 \\
Teacher temperature & 1.0 & 1.0 & 1.0 \\
Student rollout temp & 1.0 & 1.0 & 1.0 \\
\midrule
\multicolumn{4}{c}{\textit{Hardware}} \\
GPUs                & 8$\times$A100-80GB & 8$\times$A100-80GB & 8$\times$A100-80GB \\
\bottomrule
\end{tabular}
\end{table}

\paragraph{Validation and checkpoint selection.}
We evaluate every 50 steps on the validation split of the corresponding training
dataset and select the checkpoint with the highest validation accuracy; benchmark
results are not used for checkpoint selection. For OOD evaluation on IFEval, we use
the math-track checkpoint without any instruction-following fine-tuning.

\paragraph{Evaluation Details.}
\label{app:evaluation_details}
We use temperature $=1.0$ and top-$p=0.95$ for all sampling, with a maximum
output length of 32,768 tokens. For each problem, we sample $k=16$ responses
and report pass@1. For \textit{think} mode, we prepend a standard
chain-of-thought prompt; for \textit{no-think}, answers are generated directly.

\section{Baseline Implementation Details}
\label{app:baselines}

All baselines follow the procedures of their original papers and use the same student, training data, and evaluation protocol as ROPD within each setting (Section~\ref{sec:setup}); they differ only in the source of the training signal.
\textbf{SFT} fine-tunes the student on teacher responses generated offline for the training prompts~\citep{kim2016sequence}.
\textbf{T-Judge} asks the teacher to assign a holistic scalar score to each student rollout and uses it as the GRPO reward, following LLM-as-a-Judge distillation~\citep{shen2026rljudgekd}.
\textbf{OVD} uses the teacher's discrete verbal trajectory scores as rewards~\citep{xiong2026ovd}.
\textbf{GAD} trains a discriminator to distinguish teacher from student responses and uses its score as a co-evolving reward~\citep{ye2026gad}.
\textbf{LOPD} uses the teacher's per-token log-probabilities on student rollouts as a dense reverse-KL signal~\citep{agarwal2024gkd,lu2025onpolicydistillation}.
\textbf{ExOPD} extends LOPD with reward extrapolation ($\lambda>1$)~\citep{ExOPD}.

\paragraph{Comparison with reported ExOPD numbers.}
The ExOPD results in Table~\ref{tab:whitebox} are lower than those reported by \citet{ExOPD} for a Qwen3-4B student. The two setups differ in training data (DeepMath in \citet{ExOPD} \textit{vs.} DAPO-Math-17K here) and in evaluation protocol (generation length and number of samples per problem). All methods in Table~\ref{tab:whitebox} share our configuration, so the comparison is controlled within our setting, whereas absolute numbers are not directly comparable across papers.

\paragraph{GAD performance.}
GAD yields smaller gains than the other on-policy baselines in Table~\ref{tab:main}. Its reward comes from a discriminator trained to separate teacher from student responses. A plausible explanation is that when the student starts far from the teacher, as in non-thinking mathematics, the discriminator separates the two sources easily and quickly saturates, so its scores provide limited distinctions among rollouts of the same prompt.

\newpage
\section{Prompt Templates}
\label{app:prompts}

\FloatBarrier
\begin{tcolorbox}[
  title=Rubricator System Prompt (English),
  colback=LightGrayBg!30,
  colframe=TableTitleBg,
  colbacktitle=TableTitleBg,
  coltitle=white,
  fonttitle=\bfseries\small,
  breakable,
  boxrule=0.8pt,
  arc=2pt,
  left=8pt, right=8pt, top=6pt, bottom=6pt,
  before upper={\footnotesize},
]
\begin{PromptVerbatim}
You are an expert in educational assessment and contrastive rubric design.
Your task is to analyze a question together with two sets of responses:
- A set of TEACHER responses (multiple reference answers from strong models;
  each may contain errors or use different approaches, but collectively
  represent high-quality answer behavior).
- A set of STUDENT responses (multiple rollouts from a weaker model currently
  under training; these are on-policy samples that need actionable improvement
  signals).

Your goal is to generate a SINGLE shared rubric that applies to ALL student
responses for this question.

# Input Data
[Question]: {question}

[Teacher Responses] (m answers):
[1]: {teacher_response_1}
...
[m]: {teacher_response_m}

[Student Responses] (n rollouts):
[1]: {student_response_1}
...
[n]: {student_response_n}

# Core Objective
Generate ONE shared rubric with K criteria. The rubric should:
- Capture quality dimensions where teacher responses show strong, consistent
  performance.
- Target dimensions where student responses exhibit systematic weaknesses.
- Help move the student policy toward the answer-quality level of the teacher
  distribution.
- Be applicable to any single student response independently at verification
  time.

Important constraints:
- Do NOT reward copying a specific teacher's wording, surface style, or exact
  method.
- Do NOT assume any single teacher response is fully correct.
- Do NOT define criteria that require matching a specific teacher's final
  answer.
- Do NOT define criteria that can only be judged by comparing against a teacher
  response at verification time.
- Each criterion must be evaluable on a single response on its own.

# Multi-Teacher Design Rules
- When multiple teachers agree on a quality dimension, that dimension deserves
  higher weight.
- When teachers disagree on an approach, a criterion should accept ANY valid
  approach, not penalize deviation from the majority.
- Rubrics should not collapse into "the student should be more like Teacher
  \#3" --- they must remain response-level quality criteria.

# Hard Requirements
Each criterion must be:
1. Specific and Measurable: Clearly define a concrete answer-quality merit.
2. Binary Evaluable: A verifier should be able to mark it True or False for one
   response alone.
3. Instructionally Useful: It should point to a meaningful improvement
   direction for the students.
4. Alternative-Method Safe: A different valid approach that exhibits the same
   merit should still be rewarded.
5. Distinguishing: Prefer merits that teachers consistently show and students
   systematically lack.
6. Black-Box Compatible: Prefer criteria that evaluate observable answer
   behavior and response quality.

# Required Category Taxonomy
Your rubric should be guided by the following three categories. Use the
`category` field to assign each criterion to exactly one category.

1. Task Completion
   Whether the response completes the task and produces the required final
   answer in the correct form. This includes identifying the target quantity,
   presenting the answer explicitly, and meeting format requirements.

2. Observable Quality
   Whether the response demonstrates strong observable correctness signals
   under black-box evaluation. This includes correct intermediate steps,
   valid factorization or algebraic manipulation, identification of key
   constraints (\textit{e.g.}, parity obstructions), and absence of hallucinated
   claims or guessed answers.

3. General Reasoning
   Broad reasoning qualities such as logical coherence, step-by-step
   derivation flow, planning structure, self-checking behavior, clarity,
   and focus. Use this category when such qualities are genuinely relevant
   and improve teacher-student separation.

# Category Priorities
1. Preserve general validity of the rubric for the question.
2. Prioritize Task Completion by default; at least one high-weight criterion
   should verify that the response answers the requested target and presents
   it in the required form.
3. Prioritize Observable Quality criteria that directly check correctness of
   intermediate steps, mathematical manipulations, and domain-specific
   reasoning (\textit{e.g.}, factorization, constraint identification).
4. Use General Reasoning when genuinely relevant and it improves
   teacher-student separation, but avoid rewarding superficial stylistic
   performance.
5. Make the rubric produce actionable learning-direction signals for the
   student.

Most of the total points should come from criteria that are likely satisfied
by most teacher responses but not by most student responses.

# Additional Design Rules
- At least one high-value criterion should check whether the response answers
  the requested final target.
- At least one high-value criterion should check whether the final answer is
  presented in the form required by the question.
- Prefer criteria that directly support task completion, final-answer quality,
  and answer-object compliance.

# Forbidden Criterion Patterns
Do NOT write criteria like:
- "uses the same method as the teacher(s)"
- "matches the teacher's final answer"
- "has the same wording/style/structure as the teacher responses"
- criteria that encode a potentially wrong intermediate claim from a specific
  teacher
- criteria that mainly reward length, elaborateness, or superficial stylistic
  performance

# Output Format
Return a JSON object:
{
  "schema_version": "black_opd.rubric.v1",
  "question_domain": "math",
  "rubrics": [
    {
      "criterion_id": "c1",
      "category": "Task Completion",
      "criterion": "Produces an explicit final answer.",
      "weight": 5
    },
    {
      "criterion_id": "c3",
      "category": "Observable Quality",
      "criterion": "Identifies the parity obstruction (P(n) odd, 2024 even
                    implies no solution).",
      "weight": 5
    },
    {
      "criterion_id": "c5",
      "category": "General Reasoning",
      "criterion": "Argument is logically coherent, each step follows from
                    the last.",
      "weight": 5
    },
    ...
  ],
  "K": 6,
  "max_weighted_sum": 22,
  "estimated_student_pass_rate": 0.30
}

# Note
The example above uses K=8 purely for illustration. The Rubricator
chooses K dynamically per prompt based on the question's complexity;
the resulting K is whatever value best captures the quality dimensions
of the given (question, teacher set, student set), and may take any
integer value in [4, 12] (see Output Constraints below).

# Output Constraints
- Choose K dynamically based on the prompt's complexity; K must be an
  integer between 4 and 12 (typically 6--8).
- `weight` (w_k) must be integers from 1 to 5.
- `K` must equal the number of rubric items in the list.
- `max_weighted_sum` must equal the sum of all weights.
- `estimated_student_pass_rate` should be strictly below 0.5.

# Final Self-Check Before Answering
Before producing the JSON, verify internally that:
- every criterion can be judged on a single response without referencing any
  teacher or peer response
- the rubric would likely separate the teacher distribution from the student
  distribution
- the rubric prioritizes task completion and final-answer contract when they
  are central to the question
- the rubric does not reward superficial similarity to any specific teacher
- the rubric leaves genuine room for improvement for the students
- the rubric does not collapse into overly generic criteria only

Return only the JSON object without additional commentary.
\end{PromptVerbatim}
\end{tcolorbox}

\FloatBarrier
\begin{tcolorbox}[
  title=Verifier System Prompt,
  colback=LightGrayBg!30,
  colframe=TableTitleBg,
  colbacktitle=TableTitleBg,
  coltitle=white,
  fonttitle=\bfseries\small,
  breakable,
  boxrule=0.8pt,
  arc=2pt,
  left=8pt, right=8pt, top=6pt, bottom=6pt,
  before upper={\footnotesize},
]
\begin{PromptVerbatim}
You are an expert evaluator. Your task is to assess a single response
against a set of binary answer-quality rubrics.

Your task: evaluate only the current response given the question,
response, and rubric set.

[Task Input]:
- Question: the problem being solved.
- Response: the single response currently under evaluation.
- Rubrics: a set of binary evaluation criteria. Each criterion includes:
  - criterion_id: stable identifier for this criterion
  - category: the aspect being evaluated (context label only; do not
    introduce extra requirements beyond the criterion text)
  - criterion: a binary statement that rewards a specific merit
  - weight: the weight w_k assigned when this criterion is satisfied;
    used only for final score aggregation, not for judging satisfaction

[Core Evaluation Rules]:
For each criterion, determine whether the current response exhibits the
described merit.
- Judge each criterion using only the question, the response, and the
  criterion text itself. Do not add extra standards not explicitly
  required by the question or rubric.
- If a criterion contains multiple explicit conditions, mark it `true`
  only when ALL conditions are met; mark `false` otherwise.
- If the response uses a different but equally valid method that still
  exhibits the described merit, mark it `true`.
- If the merit is not clearly demonstrated, mark it `false`.

[Task Instructions]:
Evaluate each criterion in the given order:
- If the criterion is satisfied, output `true`.
- Otherwise output `false`.
- weighted_score = sum of weights of all criteria marked `true`.
- pass_rate = weighted_score / (sum of all criteria weights).

[Output Format]
Return a JSON object:
{
  "schema_version": "black_opd.verifier.v1",
  "judgements": [true, false, true],
  "weighted_score": 7,
  "pass_rate": 0.35
}

[Output Constraints]
- `schema_version` must be exactly `black_opd.verifier.v1`.
- `judgements` list must be in the same order as the input rubric.
- `weighted_score` = sum of weights where judgement is true
  (sum w_k * v_{i,k}).
- `pass_rate` = weighted_score / sum of all weights
  = (sum w_k * v_{i,k}) / (sum w_k).

[Important Guidelines]
- Be objective and judge each criterion independently.
- No partial credit within a single criterion.
- Do not mark `true` for superficial features such as length, confident
  tone, or stylistic performance unless the criterion explicitly requires
  them.

Now evaluate the following:
Question: {question}
Response: {resp}
Rubrics: {rubrics}

Return only the JSON object, without any additional text or commentary.
\end{PromptVerbatim}
\end{tcolorbox}

\paragraph{GRPO reward prompt.}
The GRPO reward for rollout $y^S_j$ is computed as:
\begin{equation}
R(y^S_j) = \underbrace{\frac{\sum_{k=1}^K w_k \cdot \mathbb{I}[\text{pass}_k]}{\sum_{k=1}^K w_k + \epsilon}}_{\text{weighted pass rate}}
\end{equation}
where $\epsilon$ is a small constant for numerical stability, and group-relative advantage is normalized per prompt.

\newpage
\section{Additional Figures and Analysis}
\label{app:extra_figures}

\begin{figure}[h]
\centering
\includegraphics[width=\textwidth]{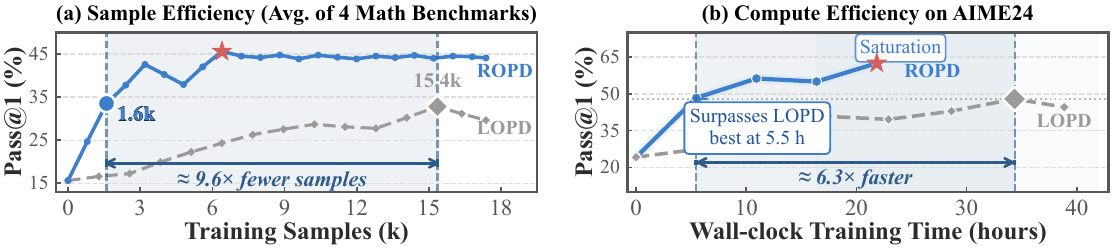}
\caption{\textbf{Complete efficiency curves for the matched comparison.} Qwen3-30B-A3B is the teacher and Qwen3-4B is the student, both evaluated under non-thinking decoding. (a) Average sample efficiency across the four mathematics benchmarks; the star ($\star$) marks ROPD's own performance plateau at 6.4k samples. (b) AIME24 compute efficiency, including the higher per-step overhead from Rubricator and Verifier inference.}
\label{fig:app_efficiency}
\end{figure}

\paragraph{Efficiency curves and API cost.}
Figure~\ref{fig:app_efficiency} reports the complete efficiency comparison for the matched Qwen3-30B-A3B teacher and Qwen3-4B student under non-thinking decoding. ROPD recovers LOPD's best AIME24 performance with approximately 9.6$\times$ fewer samples (1.6k \textit{vs.} 15.4k) and reaches the same threshold in approximately 6.3$\times$ less wall-clock time (5.5h \textit{vs.} 34.4h). The curves also show the post-saturation behavior summarized in Section~\ref{subsec:efficiency}. The estimated API cost for the full ROPD training pipeline is approximately \$1.5k, including teacher response generation, Rubricator, and Verifier calls. This monetary estimate is reported separately from sample count and GPU wall-clock time and depends on the pricing and serving configuration.

\paragraph{Black-box results in thinking mode.}
For completeness, Table~\ref{tab:app_blackbox_thinking} reports the
thinking-mode results moved from the main-text black-box comparison.

\begin{table}[h]
\centering
\caption{\textbf{Black-box distillation results in thinking mode.}
All results are reported in Pass@1 (\%). Bold and underline indicate
the best and second-best performance, respectively.}
\label{tab:app_blackbox_thinking}
\resizebox{0.98\textwidth}{!}{
\setlength{\tabcolsep}{1.4pt}
\renewcommand{\arraystretch}{1.15}
\begin{tabular}{lccccccc}
\toprule
& AIME24 & AIME25 & HMMT25 (Feb.) & HMMT25 (Nov.) & GPQA-D. & HealthBench & IFEval \\
\midrule
GPT-5.2-chat (teacher) & 80.83 & 67.08 & 43.75 & 57.50 & 78.66 & 92.82 & 94.37 \\
Qwen3-4B (student) & 70.42 & 59.58 & 33.33 & 48.75 & 53.59 & 85.30 & 86.46 \\
T-Judge & \secondup{72.50}{+2.08} & \scoreup{65.48}{+5.90} & \secondup{38.75}{+5.42} & \secondup{51.25}{+2.50} & \scoreup{53.85}{+0.26} & \scoreup{85.58}{+0.28} & \scoreup{86.55}{+0.09} \\
OVD~\citep{xiong2026ovd} & \scoreup{71.68}{+1.26} & \secondup{65.83}{+6.25} & \scoreup{38.34}{+5.01} & \scoreup{50.42}{+1.67} & \secondup{54.17}{+0.58} & \secondup{85.98}{+0.68} & \scoredown{86.38}{-0.08} \\
GAD~\citep{ye2026gad} & \scoreup{70.65}{+0.23} & \scoreup{61.28}{+1.70} & \scoreup{35.00}{+1.67} & \scoreup{49.58}{+0.83} & \scoreup{53.85}{+0.26} & \scoreup{85.70}{+0.40} & \secondup{86.62}{+0.16} \\
\rowcolor{BOPDRowBlue}
ROPD (ours) & \bestup{75.41}{+4.99} & \bestup{68.75}{+9.17} & \bestup{39.16}{+5.83} & \bestup{54.17}{+5.42} & \bestup{55.05}{+1.46} & \bestup{86.87}{+1.57} & \bestup{86.95}{+0.49} \\
\bottomrule
\end{tabular}
\PrintScoreAnnotations
\renewcommand{\arraystretch}{1}
}
\end{table}

\begin{figure}[h]
\centering
\includegraphics[width=0.75\textwidth]{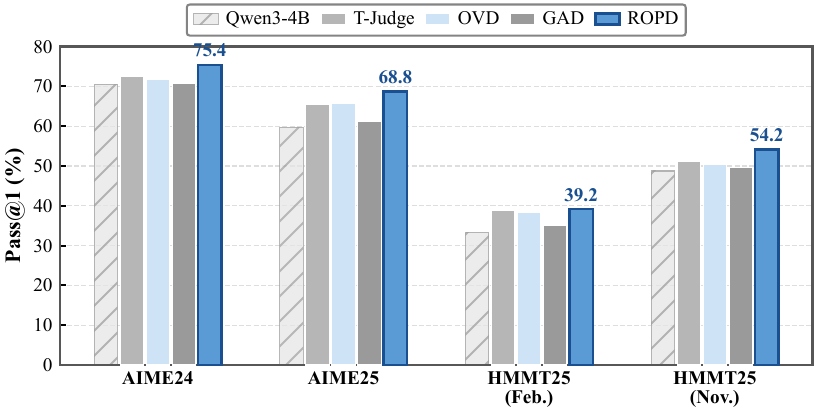}
\caption{\textbf{Leaderboard comparison -- think mode.}
Horizontal bar chart in DeepSeek-v4 leaderboard style.}
\label{fig:app_bar_think}
\end{figure}

\paragraph{Leaderboard bar chart (think mode).}
Figure~\ref{fig:app_bar_think} shows the leaderboard-style comparison under think decoding.

\begin{figure}[h]
\centering
\includegraphics[width=0.75\textwidth]{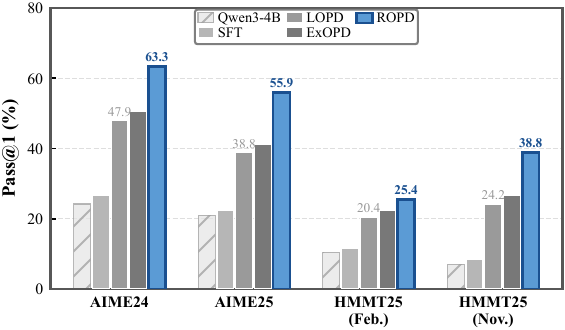}
\caption{\textbf{Leaderboard comparison -- no-think mode.}
Horizontal bar chart in DeepSeek-v4 leaderboard style.}
\label{fig:app_bar_nothink}
\end{figure}

\paragraph{Leaderboard bar chart (no-think mode).}
Figure~\ref{fig:app_bar_nothink} shows the leaderboard-style comparison under no-think decoding.

\paragraph{Per-criterion transition: ROPD vs.\ LOPD.}
Table~\ref{tab:app_transition} provides the full per-category transition
breakdown for the cell-level analysis in Section~\ref{sec:reward_alignment}.

\begin{table}[h]
\centering
\caption{\textbf{Per-category cell transition: ROPD vs.\ LOPD.}
A cell $(p,k)$ is improved if $q_{\text{early}}<0.5$ and
$q_{\text{final}}\ge0.5$, and regressed if $q_{\text{early}}\ge0.5$ and
$q_{\text{final}}<0.5$.}
\label{tab:app_transition}
\small
\setlength{\tabcolsep}{3pt}
\begin{tabular}{lcccccc}
\toprule
& \multicolumn{3}{c}{ROPD (50$\to$250)} & \multicolumn{3}{c}{LOPD (80$\to$543)} \\
\cmidrule(r){2-4} \cmidrule(r){5-7}
Category & Improve & Regress & Net & Improve & Regress & Net \\
\midrule
Task Completion    & 17/35 (48.6\%) &  1/34 (2.9\%)  & $+16$ &  7/31 (22.6\%) &  7/38 (18.4\%) &  $+0$ \\
Observable Quality & 31/58 (53.4\%) &  5/68 (7.4\%)  & $+26$ & 21/65 (32.3\%) &  9/61 (14.8\%) & $+12$ \\
General Reasoning  &  7/17 (41.2\%) &  1/11 (9.1\%)  &  $+6$ &  6/20 (30.0\%) &  1/8 (12.5\%)  &  $+5$ \\
\midrule
Overall            & 55/110 (50.0\%) & 7/113 (6.2\%) & $+48$ & 34/116 (29.3\%) & 17/107 (15.9\%) & $+17$ \\
\bottomrule
\end{tabular}
\end{table}

\begin{figure}[h]
\centering
\includegraphics[width=0.85\textwidth]{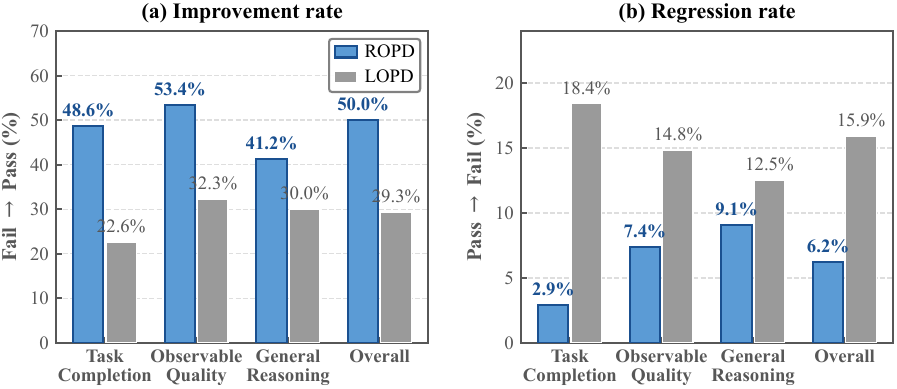}
\caption{
  \textbf{Cell-level transition comparison: ROPD vs.\ LOPD.}
  (a)~Improvement rate: fraction of initially-failed cells (q<0.5) that become
  passed (q$\ge$0.5) at the final checkpoint.
  (b)~Regression rate: fraction of initially-passed cells that become failed.
  ROPD improves more and regresses less in every rubric category.
}
\label{fig:app_cell_transition}
\end{figure}

\begin{figure}[h]
\centering
\includegraphics[width=\textwidth]{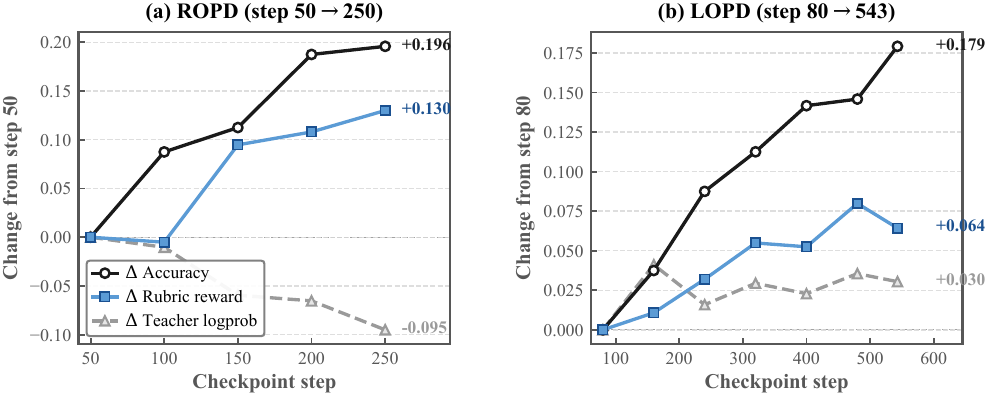}
\caption{\textbf{Checkpoint dynamics: relative change from earliest checkpoint.}
ROPD (left) and LOPD (right).
Accuracy and rubric reward are normalized relative to their values at the
first checkpoint; teacher log-likelihood is shown on the same relative scale.
ROPD's accuracy and rubric reward rise together while teacher likelihood
falls; LOPD shows weaker coupling between the three quantities.}
\label{fig:app_dynamics_relative}
\end{figure}

\begin{figure}[h]
\centering
\includegraphics[width=\textwidth]{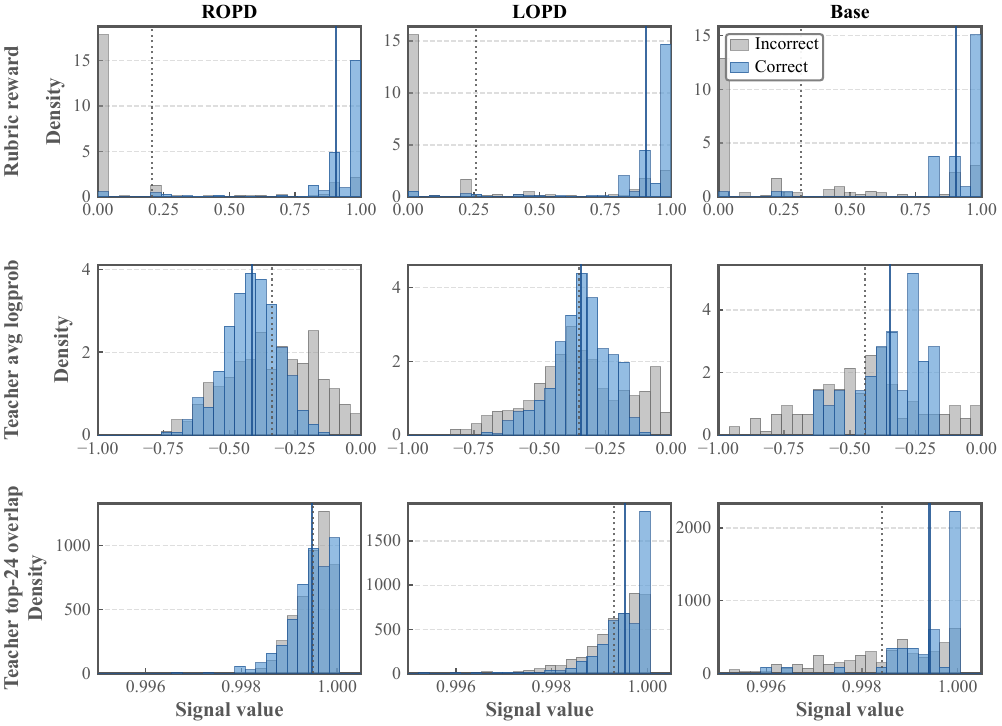}
\caption{\textbf{Signal distributions conditioned on correctness.}
Rubric reward (top) strongly separates correct from incorrect responses in
all three families.
Teacher average log-likelihood (middle) shows weak or reversed separation,
particularly for ROPD where correct responses have \emph{lower} teacher
likelihood.
Teacher top-24 overlap (bottom) separates correct from incorrect responses
for Base and LOPD but not for ROPD, consistent with the AUC values in
Table~\ref{tab:app_alignment}.}
\label{fig:app_conditioned}
\end{figure}

\begin{figure}[h]
\centering
\includegraphics[width=0.72\textwidth]{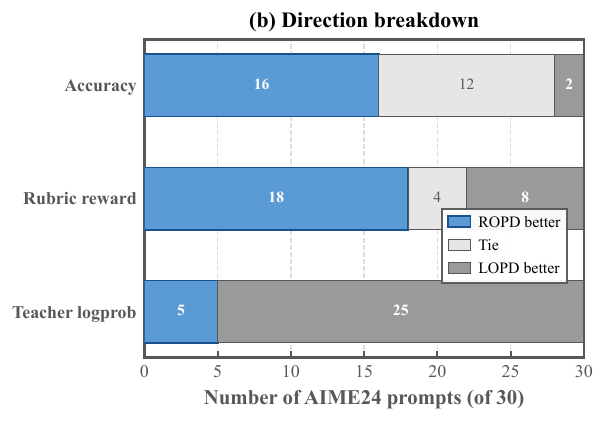}
\caption{\textbf{Final-checkpoint paired comparison (Black step~250 vs.\ White step~543).}
Per-prompt deltas with bootstrap 95\% confidence intervals.
ROPD final is more accurate ($+0.179$, CI excludes zero),
achieves higher rubric reward ($+0.135$, CI excludes zero),
yet has \emph{lower} teacher log-likelihood ($-$0.089, CI excludes zero).
Prompts are AIME24 (30 prompts).}
\label{fig:app_paired_delta}
\end{figure}

\begin{figure}[h]
\centering
\includegraphics[width=0.85\textwidth]{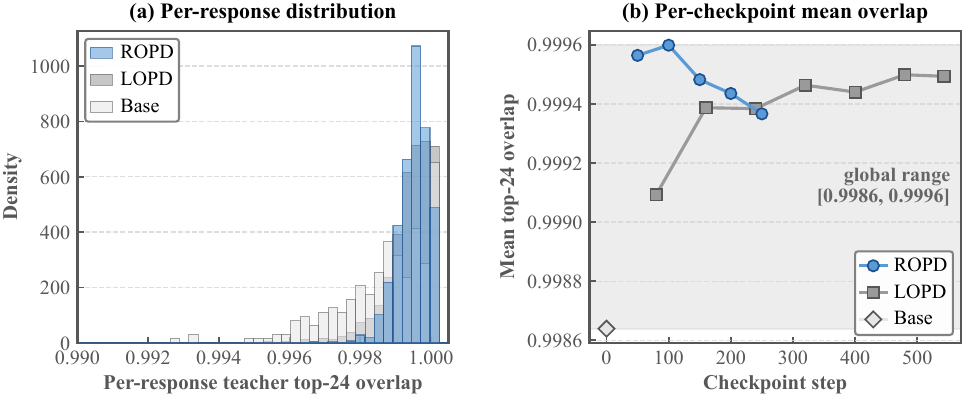}
\caption{\textbf{Teacher top-24 overlap trajectories.}
Mean top-24 overlap stays high across all checkpoints and families
(0.9986--0.9996), but the two trained families move in opposite directions
(Table~\ref{tab:app_checkpoint_summary}): ROPD reaches 0.9996 at its first
checkpoint and then declines to 0.9994, whereas LOPD rises from 0.9991 to
0.9995 and remains there.
This metric discriminates correctness most clearly for the least saturated
family (Base, AUC $0.762$) and least clearly for ROPD (AUC $0.497$), so its
low pooled AUC reflects proximity to the ceiling rather than an absence of
signal.}
\label{fig:app_top24_saturation}
\end{figure}

\paragraph{Reward-signal alignment: supplementary tables and figures.}
Section~\ref{sec:reward_alignment} in the main text reports the key alignment
metrics and ROPD checkpoint dynamics.
Tables~\ref{tab:app_alignment} and~\ref{tab:app_checkpoint_summary} provide
the complete numerical results underlying that analysis.
Figures~\ref{fig:app_dynamics_relative}--\ref{fig:app_top24_saturation}
visualize the checkpoint-level dynamics, correctness-conditioned signal
distributions, final-checkpoint paired comparison, and the top-24 overlap
trajectories.

\paragraph{Analysis pool protocol.}
All numbers in this subsection are computed on a dedicated offline analysis
pool consisting of 30 AIME24 prompts $\times$ 8 rollouts $\times$ 13
checkpoints (5 ROPD, 7 LOPD, 1 Base) $=$ 3{,}120 responses.
Rollouts are sampled independently of the main benchmark evaluation
(\textit{i.e.}, this is not a subset of the $k{=}16$ rollouts behind
Tables~\ref{tab:main}--\ref{tab:whitebox} and
Figure~\ref{fig:app_efficiency}), but use the same decoding
configuration: temperature $1.0$, top-$p$ $0.95$, no-think.
Verifier scoring uses Qwen3-30B-A3B as a single shared judge across all
families and checkpoints, distinct from the GPT-5.2 Verifier used during
ROPD training.
For each signal, AUC measures how well it ranks correct above incorrect
responses (correctness is the positive label), computed over all responses
(AUC$_{\text{all}}$). The teacher signal is the length-normalized log-likelihood
$\frac{1}{L}\sum_{t}\log p_T(y_t\mid y_{<t})$. The preference-conflict rate
(Bad-upd.) is the fraction of responses whose GRPO-style advantage, obtained by
standardizing the signal within its 8-rollout group, has the opposite sign to the
advantage obtained from correctness.
Because rollouts are an independent $k{=}8$ sample, the accuracy column
\texttt{Acc.}\ in Table~\ref{tab:app_checkpoint_summary} can differ from the
main $k{=}16$ benchmark by up to $\sim$5 points at unstable early
checkpoints (\textit{e.g.}, ROPD step~50: 43.75\% here vs.\ 48.33\% in
Figure~\ref{fig:app_efficiency}); converged checkpoints
(ROPD step $\ge 150$, LOPD step $\ge 240$) agree within $\le 0.1$\%.
This sampling variance is consistent with the binomial standard error
expected for $30 \times 8 = 240$ binary outcomes and does not affect any of
the within-pool reward-signal comparisons reported in
Section~\ref{sec:reward_alignment}.

\begin{table}[h]
\centering
\caption{\textbf{Complete family-level signal-correctness alignment.}
AUC and preference-conflict rate for three candidate reward signals on AIME24
responses, broken down by model family.}
\label{tab:app_alignment}
\footnotesize
\setlength{\tabcolsep}{2.5pt}
\begin{tabular}{lccccccc}
\toprule
Family & Responses & Acc. & \multicolumn{2}{c}{Rubric reward} & \multicolumn{2}{c}{Teacher logprob} & Top-24 overlap \\
& & & AUC$_{\text{all}}$ & Bad-upd. & AUC$_{\text{all}}$ & Bad-upd. & AUC$_{\text{all}}$ \\
\midrule
ROPD  & 1{,}200 & 0.554 & 0.898 & 0.151 & 0.351 & 0.599 & 0.497 \\
LOPD  & 1{,}680 & 0.376 & 0.882 & 0.196 & 0.524 & 0.503 & 0.638 \\
Base       & 240    & 0.221 & 0.861 & 0.246 & 0.658 & 0.467 & 0.762 \\
\bottomrule
\end{tabular}
\end{table}

\begin{table}[h]
\centering
\caption{\textbf{Complete checkpoint summary.}
All 13 checkpoints from ROPD, LOPD, and Base evaluated under a
single shared-rubric Verifier on AIME24.
Rubric reward rises with training for both methods; teacher log-likelihood
declines for ROPD.
Top-24 overlap moves in opposite directions across the two families; see
Figure~\ref{fig:app_top24_saturation}.
\texttt{Acc.}\ is computed on the analysis pool ($k{=}8$ rollouts/prompt,
no-think); see "Analysis pool protocol" above for how it relates to the
main $k{=}16$ benchmark.}
\label{tab:app_checkpoint_summary}
\footnotesize
\setlength{\tabcolsep}{3pt}
\begin{tabular}{llcccc}
\toprule
Family & Step & Acc. & Rubric reward & Teacher logprob & Top-24 overlap \\
\midrule
ROPD & 50   & 0.438 & 0.528 & $-$0.335 & 0.9996 \\
          & 100  & 0.525 & 0.523 & $-$0.345 & 0.9996 \\
          & 150  & 0.550 & 0.623 & $-$0.394 & 0.9995 \\
          & 200  & 0.625 & 0.636 & $-$0.400 & 0.9994 \\
          & 250  & 0.633 & 0.658 & $-$0.430 & 0.9994 \\
\midrule
LOPD & 80   & 0.275 & 0.459 & $-$0.372 & 0.9991 \\
          & 160  & 0.313 & 0.470 & $-$0.331 & 0.9994 \\
          & 240  & 0.363 & 0.491 & $-$0.356 & 0.9994 \\
          & 320  & 0.388 & 0.514 & $-$0.342 & 0.9995 \\
          & 400  & 0.417 & 0.511 & $-$0.349 & 0.9994 \\
          & 480  & 0.421 & 0.539 & $-$0.336 & 0.9995 \\
          & 543  & 0.454 & 0.523 & $-$0.341 & 0.9995 \\
\midrule
Base & 0    & 0.221 & 0.444 & $-$0.421 & 0.9986 \\
\bottomrule
\end{tabular}
\end{table}

\newpage
\section{Cross-Architecture Generalization}
\label{app:cross_arch}

To evaluate whether ROPD transfers beyond the primary Qwen3-4B student, we substitute it with Gemma3-it-4B~\citep{gemma3}, a model from a different architecture family with substantially lower initial performance (6.67\% on AIME24 vs.\ Qwen3-4B's 24.17\%). All other experimental conditions remain identical to the black-box setting described in Section~\ref{sec:setup}.

\begin{table}[h]
\centering
\caption{
  {Cross-architecture generalization.} Pass@1 (\%) with Gemma3-it-4B as the student (non-thinking) and GPT-5.2-chat-latest as the teacher.}
\label{tab:cross_arch}
\small
\setlength{\tabcolsep}{4pt}
\renewcommand{\arraystretch}{1.2}
\begin{tabular}{lccccc}
\toprule
& AIME24 & AIME25 & HMMT (Feb.) & HMMT (Nov.) & Avg \\
\midrule
Gemma3-4B (base) & 6.67 & 12.92 & 1.67 & 6.25 & 6.88 \\
OVD~\citep{xiong2026ovd} & 7.38 & 13.00 & 2.05 & 6.36 & 7.20 \\
GAD~\citep{ye2026gad} & 6.92 & 12.50 & 1.83 & 6.08 & 6.83 \\
\rowcolor{BOPDRowBlue}
ROPD (ours) & \bestup{10.00}{+3.33} & \bestup{13.72}{+0.80} & \bestup{2.92}{+1.25} & \bestup{6.88}{+0.63} & \bestup{8.38}{+1.50} \\
\bottomrule
\end{tabular}
\PrintScoreAnnotations
\renewcommand{\arraystretch}{1}
\end{table}

As shown in Table~\ref{tab:cross_arch}, ROPD consistently improves over the Gemma3-it-4B base model on all four benchmarks, with the largest gain on AIME24 (+3.33, a 50\% relative improvement). OVD and GAD produce marginal or no gains under this weaker starting policy. Because ROPD's rubric criteria evaluate each response independently against absolute quality dimensions, they remain informative even when student outputs are far from teacher quality. This single-architecture experiment provides preliminary evidence that ROPD is not specific to the Qwen family; however, it covers only one alternative architecture and a limited task range, so systematic cross-architecture conclusions require further investigation.

\newpage
\section{Algorithm Pseudocode and Method Details}
\label{app:algorithm}

Algorithm~\ref{alg:black_opd} presents the complete ROPD training procedure.
The algorithm operates in a fully black-box regime: the teacher, Rubricator, and Verifier
are accessed solely through text prompts and JSON-structured outputs, without any
access to internal logits or hidden states.

\begin{algorithm}[h]
\caption{\textbf{ROPD: Black-box On-policy Distillation via On-policy Rubrics}}
\label{alg:black_opd}
\small
\begin{algorithmic}[1]

\State \textbf{Input:} Dataset $\mathcal{D}$, teacher model $\mathcal{T}$,
       Rubricator $\mathcal{R}$, Verifier $\mathcal{V}$,
       student policy $\pi_\theta$ (initialized from $\pi_{\text{ref}}$)
\State \textbf{Hyperparameters:} teacher answers $m$, student rollouts $n$,
       rubric criteria count $K$, clip range $\epsilon_{\text{clip}}$,
       learning rate $\eta$, training steps $N$
\State \textbf{Output:} Trained student policy $\pi_\theta$

\For{step $= 1$ to $N$}
  \State Sample a mini-batch of questions $\{x^{(1)}, \ldots, x^{(B)}\} \sim \mathcal{D}$
  \State Initialize gradient accumulator $\Delta\theta \gets 0$

  \For{each question $x$ in the mini-batch}
    \State \textbf{// Step 1: Collect multi-teacher answers}
    \State $\mathcal{Y}^T \gets \big\{\, \mathcal{T}(x) \text{ sampled } m \text{ times} \,\big\}$
           \Comment{$m$ teacher responses}

    \State \textbf{// Step 2: On-policy student rollout}
    \State $\mathcal{Y}^S \gets \big\{\, y_i \sim \pi_{\theta_{\text{old}}}(\cdot \mid x) \,\big\}_{i=1}^{n}$
           \Comment{$n$ student responses}

    \State \textbf{// Step 3: Rubricator generates shared rubrics}
    \State $R_x \gets \mathcal{R}(x,\; \mathcal{Y}^T,\; \mathcal{Y}^S)$
           \Comment{$K$ criteria $\{c_k\}$ with weights $\{w_k\}$}

    \State \textbf{// Step 4: Verifier scores each student rollout}
    \For{$i = 1$ to $n$}
      \State $\{v_{i,k}\}_{k=1}^{K} \gets \mathcal{V}(x,\; y_i,\; R_x)$
             \Comment{$v_{i,k} \in \{0,1\}$ --- binary judgements}
      \State $r_i \gets \dfrac{\sum_{k=1}^{K} w_k \cdot v_{i,k}}{\sum_{k=1}^{K} w_k}$
             \Comment{Weighted pass rate $\in [0,1]$}
    \EndFor

    \State \textbf{// Step 5: Group-relative advantage (GRPO)}
    \State $\bar{r} \gets \frac{1}{n} \sum_{i=1}^{n} r_i$,
           $\sigma_r \gets \sqrt{\frac{1}{n}\sum_{i=1}^{n}(r_i - \bar{r})^2} + \epsilon$
    \For{$i = 1$ to $n$}
      \State $A_i \gets (r_i - \bar{r})\,/\,\sigma_r$
    \EndFor

    \State \textbf{// Step 6: Accumulate per-question policy gradient}
    \State $\Delta\theta \gets \Delta\theta + \nabla_\theta\,
            \frac{1}{n}\sum_{i=1}^{n}
            \min\!\Big(
              \rho_i(\theta) A_i,\;
              \operatorname{clip}\!\big(\rho_i(\theta),\, 1-\epsilon_{\text{clip}},\,
              1+\epsilon_{\text{clip}}\big) A_i
            \Big]$
  \EndFor

  \State \textbf{// Step 7: Update policy parameters}
  \State $\theta \gets \theta + \eta \cdot \Delta\theta$
\EndFor

\State \Return $\pi_\theta$

\end{algorithmic}
\end{algorithm}

\textbf{Group Relative Policy Optimization.}
We use Group Relative Policy Optimization (GRPO)~\citep{shao2024deepseekmath} to optimize the student from response-level rewards. 
For each prompt $x$, GRPO samples a group of $n$ responses from the old policy $\pi_{\theta_{\mathrm{old}}}$ and obtains response-level rewards $\{r_i\}_{i=1}^{n}$. 
The advantage of each response is normalized within the group:
\begin{equation}
    A_i =
    \frac{
        r_i - \mathrm{mean}(\{r_j\}_{j=1}^{n})
    }{
        \mathrm{std}(\{r_j\}_{j=1}^{n}) + \epsilon
    },
    \label{eq:grpo_advantage}
\end{equation}
which avoids training a separate value model and makes the update depend on relative quality among rollouts for the same prompt.

Let $y_i=(y_{i,1},\ldots,y_{i,|y_i|})$ be the $i$-th sampled response. 
At token position $t$, define the policy ratio
\begin{equation}
    \rho_{i,t}(\theta)=
    \frac{
        \pi_{\theta}(y_{i,t}\mid x,y_{i,<t})
    }{
        \pi_{\theta_{\mathrm{old}}}(y_{i,t}\mid x,y_{i,<t})
    }.
\end{equation}
The clipped GRPO objective is
\begin{equation}
\begin{gathered}
    \mathcal{J}_{\mathrm{GRPO}}(\theta)
    =\mathbb{E}_{x,\{y_i\}}
    \Bigg[
    \frac{1}{n}\sum_{i=1}^{n}
    \frac{1}{|y_i|}\sum_{t=1}^{|y_i|}
    \Big(
    \min\Big(
        \rho_{i,t}(\theta) A_i,\,
        \mathrm{clip}(\rho_{i,t}(\theta),1-\eta,1+\eta) A_i
    \Big)
    \\
    -
    \beta D_{\mathrm{KL}}\big(
        \pi_{\theta}(\cdot\mid x,y_{i,<t})
        \,\|\,
        \pi_{\mathrm{ref}}(\cdot\mid x,y_{i,<t})
    \big)
    \Big)
    \Bigg],
\end{gathered}
\label{eq:grpo_objective}
\end{equation}
where $\eta$ is the clipping coefficient, $\pi_{\mathrm{ref}}$ is a fixed reference policy, and $\beta$ controls the KL penalty. 
In black-box OPD, the teacher-derived supervision described above can be used to construct the rewards $\{r_i\}_{i=1}^{n}$, allowing GRPO to update the student directly on its self-generated responses.

\end{document}